\RequirePackage{fix-cm}
\documentclass[twocolumn]{svjour3}          % twocolumn
\smartqed  % flush right qed marks, e.g. at end of proof
\usepackage[ruled,linesnumbered]{algorithm2e}
\usepackage{amsmath}
\usepackage{amssymb}
\usepackage{booktabs}       % professional-quality tables
\usepackage{mathrsfs}  
\usepackage{array}
\usepackage{bm}
\usepackage{graphicx}
\usepackage{hyperref}
\usepackage{multirow}
\usepackage{url}
\usepackage{times}
\usepackage{subcaption}
\usepackage{color}
\usepackage{pifont}

\DeclareMathOperator*{\argmin}{arg\,min}
\DeclareMathAlphabet{\mathcal}{OMS}{cmsy}{m}{s}
\newcommand{\eg}{\textit{e.g.},~}
\newcommand{\ie}{\textit{i.e.},~}
\newcommand{\etal}{\textit{et. al}~} 
\newcommand{\att}[1]{\textcolor{red}{{}#1}}

\newcommand{\cmark}{\ding{51}}
\newcommand{\xmark}{\ding{55}}
\newcommand\undermat[2]{
  \makebox[0pt][l]{$\smash{\underbrace{\phantom{
    \begin{matrix}#2\end{matrix}}}_{\text{#1}}}$}#2}
\newcommand{\colvecn}[2][.9]{%
  \scalebox{#1}{%
    \renewcommand{\arraystretch}{.9}%
    $\begin{bmatrix}#2\end{bmatrix}$%
  }
}
\newcommand{\colvecs}[2][.6]{%
  \scalebox{#1}{%
    \renewcommand{\arraystretch}{.6}%
    $\begin{bmatrix}#2\end{bmatrix}$%
  }
}
\newcommand{\colvect}[2][.5]{%
  \scalebox{#1}{%
    \renewcommand{\arraystretch}{.5}%
    $\begin{bmatrix}#2\end{bmatrix}$%
  }
}
\SetKw{KwInn}{in}
\begin{document}
\sloppy

\title{Learning Structure-Supporting Dependencies via Keypoint Interactive Transformer for General Mammal Pose Estimation}



\author{Tianyang~Xu         \and
        Jiyong~Rao          \and
        Xiaoning Song       \and
        Zhenhua~Feng       \and
        Xiao-Jun~Wu                %etc.
}


\institute{T.~Xu \at
              School of Artificial Intelligence and Computer Science\\
              Jiangnan University, Wuxi 214122, China. \\
              \email{tianyang.xu@jiangnan.edu.cn}           %  \\
           \and
           J.~Rao \at
              School of Artificial Intelligence and Computer Science\\
              Jiangnan University, Wuxi 214122, China. \\
              \email{raojiyong@stu.jiangnan.edu.cn}    
           \and
           X.~Song \at
              School of Artificial Intelligence and Computer Science\\
              Jiangnan University, Wuxi 214122, China. \\
              \email{x.song@jiangnan.edu.cn}           %  \\
           \and
           Z.~Feng \at
           School of Computer Science and Electronic Engineering\\
           University of Surrey, Guildford GU2 7XH, UK. \\
              \email{z.feng@surrey.ac.uk} 
        \and
           X.-J.~Wu \at
              School of Artificial Intelligence and Computer Science\\
              Jiangnan University, Wuxi 214122, China. \\
              \email{wu\_xiaojun@jiangnan.edu.cn}
}

\date{Received: date / Accepted: date}

\maketitle

\begin{abstract}
General mammal pose estimation is an important and challenging task in computer vision, which is essential for understanding mammal behaviour in real-world applications. However, existing studies are at their preliminary research stage, which focus on addressing the problem for only a few specific mammal species. In principle, from specific to general mammal pose estimation, the biggest issue is how to address the huge appearance and pose variances for different species. We argue that given appearance context, instance-level prior and the structural relation among keypoints can serve as complementary evidence. To this end, we propose a Keypoint Interactive Transformer (KIT) to learn instance-level structure-supporting dependencies for general mammal pose estimation. Specifically, our KITPose consists of two coupled components. The first component is to extract keypoint features and generate body part prompts. The features are supervised by a dedicated generalised heatmap regression loss (GHRL). Instead of introducing external visual/text prompts, we devise keypoints clustering to generate body part biases, aligning them with image context to generate corresponding instance-level prompts. Second, we propose a novel interactive transformer that takes feature slices as input tokens without performing spatial splitting. In addition, to enhance the capability of the KIT model, we design an adaptive weight strategy to address the imbalance issue among different keypoints. Extensive experimental results obtained on the widely used animal datasets, AP10K and AnimalKingdom, demonstrate the superiority of the proposed method over the state-of-the-art approaches. It achieves 77.9 AP on the AP10K \textit{val} set, outperforming HRFormer by 2.2. Besides, our KITPose can be directly transferred to human pose estimation with promising results, as evaluated on COCO, reflecting the merits of constructing structure-supporting architectures for general mammal pose estimation. Code is available at \url{https://github.com/Raojiyong/KITPose}
% From the perspective on learning input-wise keypoint dependencies

\keywords{Mammal Pose Estimation\and Keypoint Interactive Transformer\and Adaptive Weight Strategy\and Human Pose Estimation.}
\end{abstract}
% \fi
% \IEEEpeerreviewmaketitle
%
\section{Introduction}
\label{sec_intro}
Animal pose estimation~\cite{mathis2018deeplabcut,cao2019cross,li2021synthetic,mu2020learning,rao2022kitpose} aims to localise predefined keypoints for deformable animal instances, which has gradually attracted increasing attention in recent years.
By employing animal pose estimation, the community can obtain a better understanding so as to develop several practical visual applications, such as animal behaviours analysis~\cite{zhang2023animaltrack,pereira2022sleap}, zoology study~\cite{zuffi2019three,ng2022animal}, and wildlife conservation programs~\cite{davies2015keep}.

Currently, most pose estimation research is human-oriented.
One of the most representative approaches, High-Resolution Network (HRNet)~\cite{sun2019deep}, achieves remarkable performance in human pose estimation by maintaining the low-to-high representations and fusing the multi-scale features, with a balanced perception of context details and content semantics.
Followed by DarkPose~\cite{zhang2020distribution}, a novel heatmap decoding method, significantly improves the estimation performance without additional complex model design.
Inspired by the success of vision transformers, researchers began applying transformers to human pose estimation.
By integrating the advantages of CNN and Transformer modelling,  Transpose~\cite{yang2021transpose} and TokenPose~\cite{li2021tokenpose} employ CNN to obtain strong locality inductive bias and then utilise Transformer to capture long-range spatial information.
On the other hand, HRFormer~\cite{yuan2021hrformer} also maintains high-resolution representations and directly adopts the transformer backbone in the feature extraction stage.
Unlike the previous statement, ViTPose~\cite{xu2022vitpose,xu2022vitpose+} simply follows the network architecture of hierarchical vision transformers~\cite{dosovitskiy2020image} to construct a pure transformer network, achieving a new state-of-the-art performance record on various pose estimation benchmarks.

% Although animal pose estimation is analogous to human pose estimation to some extent, we argue that the two tasks are quite different.
Although the pipeline of human pose estimation can be directly applied to realise animal pose estimation, the network design necessitates further deliberations.
The first difference lies in the data volume present in both.
To the best of our knowledge, the largest mammal pose estimation dataset AP-10K \cite{yu2021ap}, which consists of 10K images, is still much smaller than the widely studied human pose dataset, \ie COCO~\cite{lin2014microsoft} with 200K images.
Such a small animal data collection is inadequate to train deep learning-based models with similar parameter volumes as those for humans.
As shown in Table~\ref{tab:hp_method}, the representative human pose estimators have shown degraded performance on animals.
In particular, the sophisticated transformer-based approaches produce even worse results, \eg TokenPose~\cite{li2021tokenpose} drops $3.1$ AP points.
In essence, there is a huge gap between the animal and human poses.
Specifically, animal pose estimation involves multiple species, while human pose estimation only focuses on one category.
As a result, there are large data variances for both intra- and inter-family instances.
For example, panthers and cats, both belong to the same family of Felidae, the former tends to be very large with dark skin colour, while the latter is smaller with bright skin colour.
Furthermore, for two distinct families of animals, the neck of giraffes is far longer than those of monkeys, which means that there is a large margin in the shape between the two.
The above variations extremely challenge the general mammal pose estimation compared to the intensively studied problem of human pose estimation.
\begin{table}[t]
    \renewcommand{\arraystretch}{1.2}
    % \centering
    \setlength{\tabcolsep}{1mm}
    \begin{tabular}{lcccc}
    \toprule
    Methods &   Source & Backbone    &   COCO(AP)    &   AP10K(AP)   \\
    &&&[Human]&[Animal]
    \\
    \midrule
    HRNet       & CVPR2019  &   HRNet-W32   &   74.4        &   73.8(\textcolor{green}{$\downarrow0.6$})    \\
    DarkPose    & ICCV2020  &   HRNet-W32 &   75.6        &   74.6(\textcolor{green}{$\downarrow1.0$})    \\
    TokenPose   & ICCV2021  &   L/D24       & 75.8  & 72.7(\textcolor{green}{$\downarrow3.1$})       \\
    HRFormer    & NIPS2021  & HRFormer-B    & 75.6  & 74.5(\textcolor{green}{$\downarrow1.1$})  \\
    ViTPose+    & TPAMI2023  & Vit-B     & 77.0  & 74.5(\textcolor{green}{$\downarrow2.5$}) \\
    \bottomrule
    \end{tabular}
    \caption{Performance comparisons between human and animal pose estimation.}
    \label{tab:hp_method}
\end{table}

Considering the fundamental importance of understanding animal behaviour, animal pose estimation has been separated from human pose estimation.
However, primary studies overlooked the design of a unified model for general animal pose estimation and tend to address the task for only specific categories.
For example, DeepLabCut~\cite{mathis2018deeplabcut,mathis2021pretraining} and LEAP~\cite{pereira2019fast} both propose to use transfer learning to mitigate the insufficient animal training data issue, where the models pre-trained from other vision tasks are directly employed.
The extensions of DeepLabCut~\cite{lauer2022multi} and SLEAP~\cite{pereira2022sleap} design the framework of multi-animal pose estimation and tracking, which still focuses on specific animal species, such as mouses, houses, flies and so on.
With the recently released large-scale animal datasets, \ie AP-10K~\cite{yu2021ap} and Animal Kingdom~\cite{ng2022animal}, which contain over 10K images with 54 species and 7K images with 123 mammal species, respectively, the community shifts its focus on addressing the data diversity within comparably small data volume for each species.
To further employ external information, CLAMP~\cite{zhang2023clamp} employs CLIP~\cite{radford2021learning} to provide rich prior knowledge to compensate for the lack of visual cues, introducing text prompts to assist animal pose estimation.
The few/zero-shot results show the effectiveness and generalisation of the prompt-based solutions.

To this end, this paper addresses the multiple species mammal pose estimation from a model-design perspective and reveals a generic factor for this task, \ie the structure-supporting keypoint dependencies.
In many scenarios, the structural dependencies between keypoints are similar across different animal species, \eg eyes are next to the nose and the shoulders are farther from the knees.
Specifically, after analysing the distribution of quadruped animal and human keypoints, \cite{jiang2022animal} suggested that although the species are different, the definition and distribution of the keypoints are analogous to a large extent among general mammals.
To study the structure relevance, AnimalPose~\cite{cao2019cross} identified the structural information of a certain animal category is helpful to estimate the poses of other species if they share a similar structure.
To promote the supervision of the shared structure, Ke \etal\cite{ke2018multi} proposed a structure-aware loss to force the model to learn pairwise even high-order keypoint matching patterns.
GraphCape~\cite{hirschorn2023pose} followed the CAPE~\cite{xu2022pose} paradigm, introducing the graph-based model and treating keypoints as individual nodes in the decoder part. GraphCape leverages inherent geometrical relations between keypoints to preserve structure.
Therefore, %if a pose varies heavily, 
the location of the keypoint can be inferred according to its relevant keypoints in a pre-defined structure graph.
However, the subsequent problem is that when facing significant variations in pose, a fixed graph is not always the optimal structure.
The phenomenon reflects the typical difficulty in nonhuman animal pose estimation, where large variations in poses are frequently encountered.
For example, when an antelope is standing, the correlation between its nose and eyes could be strong, but when it bends down to drink, the distance between its nose and front paw decreases, potentially increasing their correlation.
Drawing on these, identifying the intrinsic relevance among keypoints is of critical importance for detecting the keypoints for animals.

% -------------------------- Bad Activation Map & Attention Map ---------------------------------
\begin{figure}[t]
    \centering
    \begin{subfigure}{0.45\textwidth}
        \includegraphics[trim={0mm 1mm 105mm 0mm},clip,width=1\linewidth]{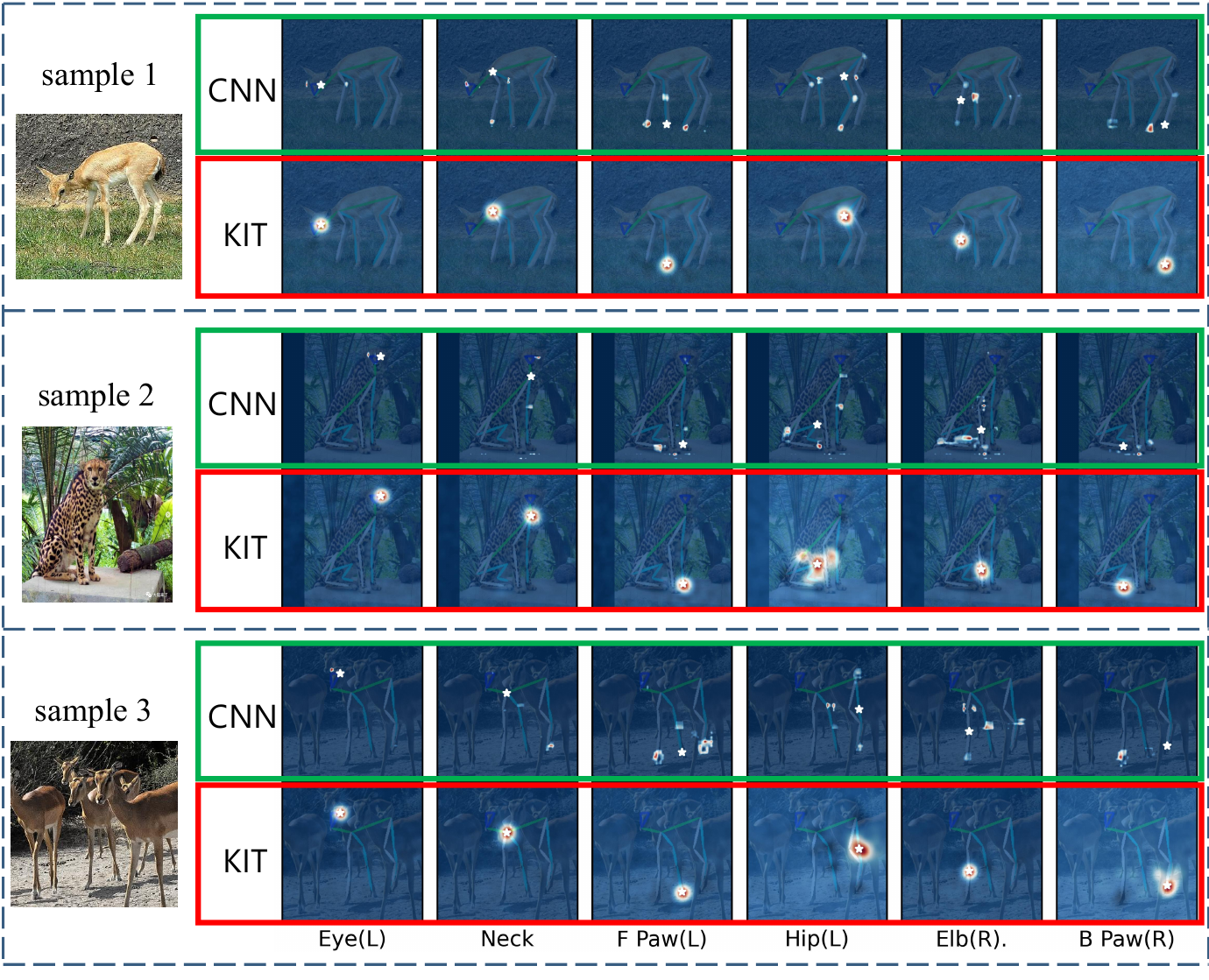} 
        \caption{Keypoint features}
    \label{fig:intro-activation}
    \end{subfigure}%

    \begin{subfigure}{0.45\textwidth}
        \includegraphics[width=\textwidth]{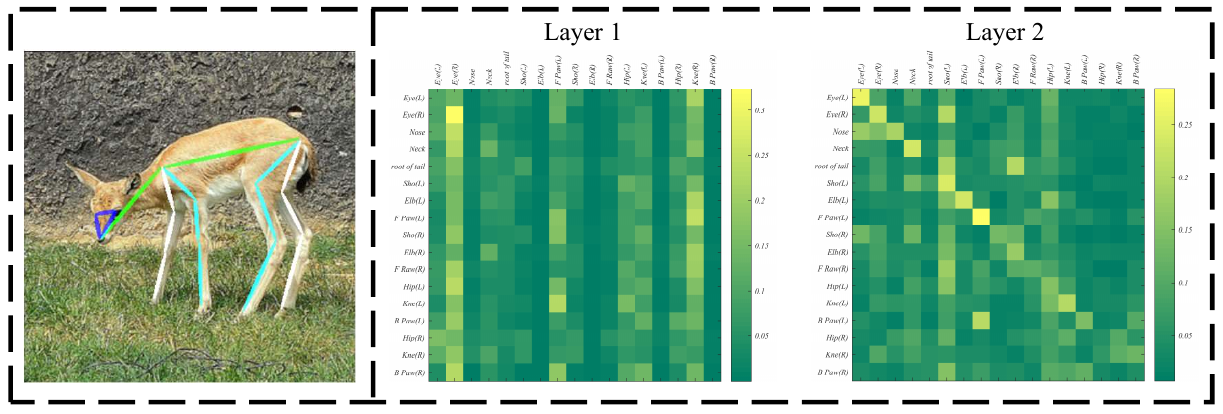}
        \caption{Attention map}
        \label{fig:intro-attention}
    \end{subfigure}
    
    % \subfigure[Keypoint features]{
    % \begin{minipage}[t]{0.45\linewidth}
    % \centering
    % \includegraphics[width=1in]{Figures/activation_map.pdf}
    % %\caption{fig1}
    % \end{minipage}%
    % }%
    % \subfigure[Attention map]{
    % \begin{minipage}[t]{0.45\linewidth}
    % \centering
    % \includegraphics[width=1in]{Figures/Intro_attn_map.pdf}
    % %\caption{fig2}
    % \end{minipage}%
    % }%
\caption{(a) The obtained feature maps from the backbone and the proposed KIT module. (b) The visualisation of the attention maps of the KIT module.}
\label{fig:activation_map}
\end{figure}
% ---------------------------------------------------------------------------------------------------

In our study, we present a novel structure-supporting architecture to obtain the dependencies among different keypoints, namely Keypoint-Interactive Transformer (KIT).
KIT investigates the potential cross-point relevance through self-attention techniques, enabling the power of inferring hard keypoints from well-detected keypoints without requiring a large amount of training data.
Specifically, in KITPose, we start from a high-resolution backbone~\cite{sun2019deep}, which is widely used as a standard feature extractor for pose estimation tasks.
After obtaining the feature maps, our KIT module performs stacked attention operations in the decision stage, achieving hierarchical interactions.
Meanwhile, in order to prevent overfitting when faced with a small training data volume, we employ the cutmix~\cite{yun2019cutmix} augmentation.
The strategy forces the network to develop a strong localisation ability via image corruption, thereby further enhancing the model's generalisation.
We present some intuitive examples of the backbone features and the coarse KIT module output in Fig.~\ref{fig:activation_map}.
It can be observed that the backbone features always produce a multi-modal heatmap rather than a uni-modal heatmap for one specific keypoint.
While after endowing our structure-supporting interactions to the backbone features, the decision for each keypoint can be concentrated on one specific location.

To obtain satisfied intermediate features, we propose a generalised heatmap regression for intermediate supervision, which can prevent the intermediate layers from producing predominant keypoint features.
Specifically, similar to the customary introduction of intermediate supervision, we also incorporate a novel supervised signal for heatmaps, namely Generalised Heatmap Regression Loss (GHRL), to enforce structural discriminability while emphasising localising precision.
In principle, it is difficult to define what discriminative keypoint features should entail.
Towards this issue, we combine two distinct transformations to the intermediate targets, one to smooth them and the other to sharpen them, which alters the static-designed heatmap and prevents different keypoints from producing uniform-shaped representations.
The smoother one contains more semantic context, while the sharper one processes more precise localisation capability.
% The manually-designed target are relatively generalizable.
On this basis, our GHRL aims at balancing the two types of targets, which enables the intermediate output to adaptively adjust suitable keypoint features for KIT.

\begin{figure*}[t]
\begin{center}
\includegraphics[trim={0mm 5mm 0mm 5mm},clip,width=1\linewidth]{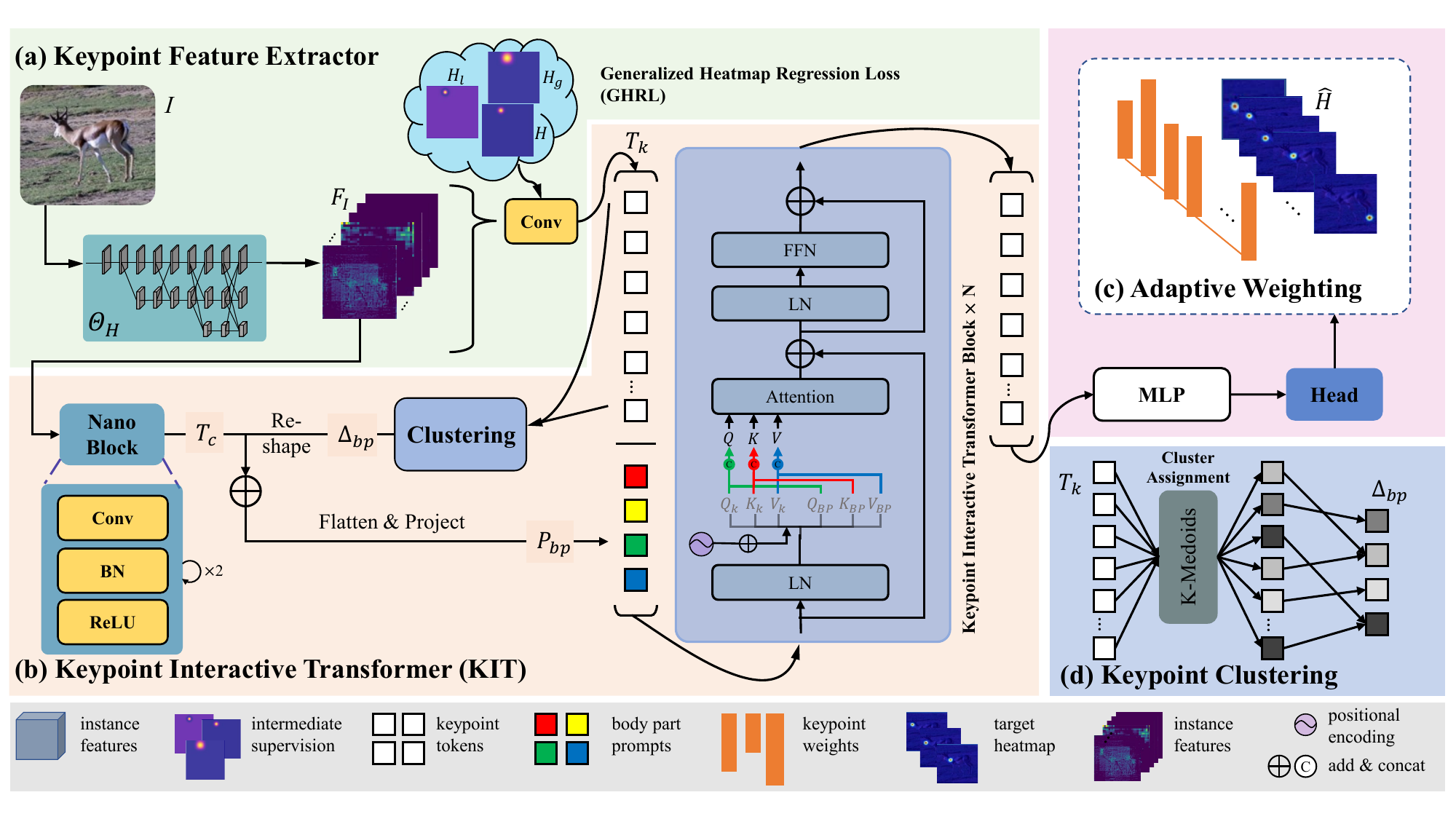}
\end{center}
\caption{Overview of KITPose. Firstly, the instance features are extracted by the backbone. In the keypoint branch, the GHRL loss function is used as an intermediate supervision. Then the features are directly flattened to 1D tokens along the channel dimension. In the context branch, the NanoBlock is utilised to extract instance-level context tokens. The context tokens are then fused with the body part biases obtained through clustering of keypoint tokens to generate body part prompts. The structural relationship and constraint are captured through self-attention interactions in each Transformer block. Finally, an output head is used to predict the keypoints heatmaps.}
\vspace{-1em}
\label{fig:pipeline}
\end{figure*}

Furthermore, given the substantial variations among different animal instances, we introduce a novel instance-level adaptation: \textit{body part prompt}.
The key idea involves generating priors for body parts based on each input instance and subsequently integrating them with the image context to produce the body part prompts.
We first cluster keypoint features to serve as prior knowledge for the body parts of different instances.
Besides, we extend the feature extractor by further learning a lightweight network to generate instance-level context tokens.
The overall architecture of KITPose is illustrated in Fig.~\ref{fig:pipeline}.
It should be noted that the Transformer employed in our design takes feature map slices as input tokens, which differs from the existing spatial units arrangement.
Involving the instance- and keypoint-level interactions enables comprehensive decision-level fusion, explicitly reflecting the point-to-point supportive power.

In terms of the keypoint-wise importance, we find that existing pose estimation methods~\cite{li2021tokenpose,yuan2021hrformer,xu2022vitpose,sun2019deep,xiao2018simple} pre-define the keypoint weights manually, resulting in degraded performance when transferring to other datasets.
We explore three strategies to set keypoint weight, \ie hand-crafted, constrained, and adaptive manners.
The hand-crafted keypoint weights are selected by experience preference or experimental evaluation.
In general, the keypoints, at different body parts, are semantically discriminative.
Therefore, in addition to leveraging hierarchical interactions among keypoints, it is essential to consider the individual importance of each specific keypoint.
A common practice is to manually set a scaling factor before training, where the scaling factor is constrained and used to re-weight losses for different keypoints.
Meanwhile, the same keypoint in different instances could also lead to distinct effects on pose estimation across multiple species.
In other words, specific keypoints should be treated differently.
Inspired by the design of FocalLoss and its variants~\cite{lin2017focal,li2020generalized} in object detection tasks, which overcomes the class imbalance problem, we propose an adaptive keypoint weights map, which adaptively down-weight the loss of well-detected samples, guiding the model to focus on the harder keypoints.
Therefore, this manner can balance the instance- and keypoint-specific importance.

In summary, our contributions are as follows:
\begin{itemize}
    \item We introduce a novel body part prompt scheme to handle large pose variances of general mammals. The body parts employ context distributions from the feature extractor to provide richer instance-specific semantic context. 
    % More comprehensive visual cues can be obtained to better perform the keypoint interactions.
    %The prompts are generated before the keypoint interaction to be better guided by such instance-level semantic context.}
    \item A conceptually simple but highly effective architecture Keypoint-Interactive Transformer is proposed to emphasise instance-level interactions between keypoints and body parts, achieving structure-aware pose estimation. With the learned specific relationships, the model exhibits significant robustness and generalisation on cross-species pose estimation.
    \item We propose an adaptive keypoints weight strategy to replace the fixed-designed keypoint weighting strategy. Keypoints importance can be flexibly reflected in our adaptive mechanism, delivering improved performance.
    \item Extensive evaluation is conducted on the predominant benchmarks, \ie AP10K~\cite{yu2021ap}, Animal Kingdom \cite{ng2022animal}, and COCO~\cite{lin2014microsoft}. The experimental results demonstrate the superior performance of the proposed KITPose method over the state-of-the-art, in terms of both effectiveness and robustness.
    % The proposed KITPose-E2C4 with HRNet-W48 backbone achieves 77.9 AP, 78.4 AP and 77.3 AP on AP10k \textit{val}, COCO \textit{val} and \textit{test-dev} sets respectively.
    It obtains competitive SOTA results on the MS COCO dataset without the use of elaborate architecture designs or complex training strategies.
    
\end{itemize}

% Overall model architecture, KITPose

% -------------------------------------------------------Related Work-----------------------------------------------------------------
\section{Related Work}
\label{sec_2}
\subsection{2D mammal pose estimation}
\label{subsec:rel_pe}
Animal pose estimation, similar to human pose estimation, is a fundamental task in computer vision and refers to localising all body keypoints for animal-targeted images.
Existing studies have circumvented the proposition of effective models and, instead addressed the problem from the perspectives of transfer learning and domain generalisation.
Specifically, AnimalPose~\cite{cao2019cross} leverages a cross-domain adaptation scheme to learn pose similarities between animals and humans.
To cope with the limited labelled animal data volume, direct attempts~\cite{mu2020learning,li2021synthetic} learn prior-aware knowledge from synthetic animal datasets, and generate pseudo labels for unlabelled data.
In terms of refining the pseudo labels, ScarceNet~\cite{li2023scarcenet} proposes a novel semi-supervised learning method that improves the reliability of pseudo label generation, effectively mitigating the data scarcity problem.
Similarly, CLAMP~\cite{zhang2023clamp} guides the learning process by incorporating external textual information, with promising performance achieved in both supervised and few-shot settings.
% \att{UniAP~\cite{sun2024uniap} proposes a multi-task learning framework to address universal animal perception task.
% It takes multi-modal labels in the support sets as prompts to guide leaning, which achieves impressive performance even in few-shot setting.}
However, there have been few efforts to explore the potential of effective model structures without introducing external data.

This paper introduces the keypoint dependencies by using keypoint and prompt attention.
We note that the appearance and pose may vary heavily across mammal species, and the intrinsic dependencies among different keypoints also change with varying instances accordingly.
With this insight, we propose a novel Keypoint-Interactive Transformer for effectively learning instance-level structure-supporting keypoint dependencies.

\subsection{Vision Transformers}
\label{subec:vit}
The success of Vision Transformers (ViTs)~\cite{dosovitskiy2020image,liu2021swin} on many vision tasks has been undisputed in recent years.
Given their capacity for establishing long-range dependencies, some studies~\cite{park2022how,naseer2021intriguing,mao2022towards} observed that SAs exhibit intriguing properties compared to CNN approaches.
In particular, performing SAs for vision tasks produces improved robustness against image occlusions, perturbations, and domain shifts.
The underlying reason lies in that SAs achieve generalised spatial smoothing, which means that SAs obtain an appropriate inductive bias during training.
Recently, many pose estimation researchers~\cite{yang2021transpose,li2021tokenpose,yuan2021hrformer,xu2022vitpose,li2021prtr} are intrigued by the advantages of SAs.
They attempt to involve Transformer architecture in pose estimation and deliver promising performance.
For instance, HRFormer~\cite{yuan2021hrformer} learns high-resolution representations along with the local-window self-attention module, decreasing the time consumption of the standard global-window formulation.
Similarly, TransPose~\cite{yang2021transpose} directly employs the Transformer to learn global dependencies on the features extracted by CNN.
TokenPose~\cite{li2021tokenpose} introduces external tokens to represent keypoints and interact with the CNN image features.
ViTPose~\cite{xu2022vitpose}, without using CNN backbones, designs a simple yet effective plain Vision Transformer architecture, achieving a new state-of-the-art in human pose estimation benchmarks.
% However, these approaches tend to get inferior performance in animal pose estimation, as the current data volume for animal pose estimation cannot support sufficient training, not to mention the large data variances across multiple mammal species.
However, these approaches tend to get inferior performance in animal pose estimation, as the limited size of current animal datasets is insufficient to support adequate training.
Besides, the Transformer-based methods have not taken into account the significant variations among different animal instances.
Drawing on this, we propose to explicitly introduce instance-level information and interact with keypoint tokens.

\section{Keypoint-Interactive Transformer}
\label{sec_3}
In our design, we formulate the mammal pose estimation solution following a top-down scheme with the heatmap regression paradigm.
A mammal instance is first cropped using the annotated bounding box.
Then, the proposed KITPose network outputs the locations of the involved keypoints within the instance.
As shown in Fig.~\ref{fig:pipeline}, KITPose exploits a hybrid architecture of the convolutional blocks and self-attention structures.
In terms of the convolution part, we choose HRNet as the backbone.
Particularly, we introduce a novel GHRL loss function to supervise the backbone features, thereby enhancing their discriminative capabilities.
After obtaining the backbone features, we cluster the instance-level visual prompts and subsequently rebuild the transformer blocks to drive instance-level structure-supporting interactions among keypoints.
The transformer encoder consists of several Keypoint Interactive Transformer (KIT) layers in a stacked manner.
The details of the architecture are described as follows.

\subsection{High-Resolution Backbone}
\label{subsec:hrnet}
High-resolution representations play a critical role in performing fine-grained prediction~\cite{yuan2021hrformer,sun2019deep,xiao2018simple}, which precisely satisfies the demand of pose estimation.
The success can be attributed to two main aspects, \textit{i.e.}, high-resolution preservation with high-to-low-resolution mapping, and multi-granularity fusion via repeated cross-resolution passing.
We, therefore, directly leverage HRNet~\cite{sun2019deep} to extract the keypoint feature representations.
For the input image $I\in\mathbb{R}^{3\times H\times W}$, the feature extractor $\Theta_H$ is firstly utilised to extract the instance features $F_I=(\Theta_H(I))$, followed by transforming to keypoint features $F_k=\text{Conv}(F_I)$, where Conv denotes an $1\times1$ convolution to adjust the numbers of channels.
$F_k$ is then flattened and linearly transformed to yield keypoint tokens $T_k\in\mathbb{R}^{N\times C}$, where $N$ denotes the number of keypoints and $C$ is the hidden dimension.

To obtain satisfied intermediate keypoint representations, we propose the Generalised Heatmap Regression Loss (GHRL) to represent the general distribution of keypoint locations instead of a fixed Gaussian shape.
Consequently, we can obtain more suitable and reliable intermediate representations, while gaining insight into their various potential distributions.
The formulation of GHRL can be written as:
\begin{equation}
\begin{aligned}
\label{equ-ghrl}
    \text{GHRL}(F_k,H)=&\lvert F_k-H\rvert^\beta\Bigl( \lvert H_l-F_k\rvert\mathcal{L}_{mse}(F_k,H_l) \\
    &+\lvert H_g-F_k\rvert\mathcal{L}_{mse}(F_k,H_g)\Bigr)
\end{aligned}
\end{equation}
where $H$ are the ground-truth heatmaps.
$H_l$ refers to the heatmaps processed through Laplacian filtering with a standard $3\times 3$ kernel,
% a kernel size of [[0, -1, 0], [-1, 4, -1], [0, -1, 0]], 
while $H_g$ refers to the heatmaps processed through a Gaussian filter with a kernel size of 13 and the sigma set to 4.
Intuitively, $H_l$ forces prediction to rapidly focus on the locations of the keypoint, and $H_g$ aims at loosening the constraint to allow more tolerated shifts.
Based on Eq.~\eqref{equ-ghrl}, different keypoint features can adaptively reflect the varying roles played in subsequent interactions.

\subsection{Body Part Prompts}
\label{body_part_prompt}
To refine the model capacity with learnable prompt tokens, TokenPose~\cite{li2021tokenpose} and CLAMP~\cite{zhang2023clamp} use additional tokens to represent and interact with keypoints.
However, the former approach lacks consideration of the diversity of keypoints among different instances, while the latter approach necessitates the usage of CLIP~\cite{radford2021learning} as an external feature provider.
We perform clustering on keypoint tokens $T_k$ of a certain animal instance to generate visual body part prompts $P_{\text{bp}}\in\mathbb{R}^{N_p\times C}$ accordingly, where $N_p$ denotes the number of prompts.
This is based on the assumption that correlated keypoints are more likely to possess similar features. At the same time, these keypoint tokens can explicitly form body parts to provide more semantic and discriminative information.
Subsequently, we further learn a lightweight network to generate the context tokens $T_c\in\mathbb{R}^{N_p\times C}$ for each input instance, which is then fused with the body part biases $\Delta_{\text{bp}}\in\mathbb{R}^{N_p\times C}$ to form instance-level $P_{\text{bp}}$.
See Fig~\ref{fig:pipeline} for an illustration of the pipeline.

\textbf{Token Clustering.}
Motivated by CenterClip~\cite{zhao2022centerclip}, we employ k-medoids++, a variation of k-means, as our clustering solution.
The algorithm detail is shown in Algorithm~\ref{algo-cluster}.
Given a set of keypoint tokens $T_k$ and the number of cluster $N_p$, the goal of the clustering stage is to partition the $N$ observations into $N_p$ representations relevant to the current instance, \ie $M=\{\phi_1,\phi_2,\cdots,\phi_{N_p}\}$.
Following the heuristic initialisation method KKZ~\cite{katsavounidis1994new} (as expounded in CenterClip~\cite{zhao2022centerclip} for its ability to enhance medoids discrimination and accelerate convergence), we select the point with the highest $\ell 2$-norm as the first medoid, followed by selecting the farthest point from the existing medoids as the second medoid.
% We take the Euclidean distance as the dissimilarity measure for efficient calculation. 
Thus, the objective of clustering is to obtain an assignment that minimises the within-cluster sum of Euclidean distances.
According to the assignment, we obtain the body part biases $\Delta_{\text{bp}}$ by calculating the mean of the sample points belonging to cluster $\mathcal{C}$, respectively.

\begin{algorithm}[t]
	\caption{The K-Medoids++}
	\label{algo-cluster}
	\KwIn{A instance keypoint set $T_k=\{k_{1}, ... , k_{N} \}\in\mathbb{R}^{N\times C}$, cluster number $N_p$}
%	\KwData{current period $t$, initial inventory $I_{t-1}$, initial capital $B_{t-1}$, demand samples}
	\KwOut{cluster assignments $\mathcal{C}$ and final medoids $\mathcal{M}$}
 	% $\phi_1 \leftarrow \argmax_{x_i} \lVert x_i \rVert_2$\;
        $\mathcal{M} \leftarrow$ KKZ initialisation method\;
 	\While{Not converged}{
            $\mathcal{M}_{\text{prev}}\leftarrow \mathcal{M}$\;
            \For{$\phi_j$ \KwInn $\mathcal{M}$}{
                $\mathcal{C}_j=\varnothing$\;
                \For{$k_i$ \KwInn $T_k$}{
                    % \tcp{$\phi_j$ is the mean of the points in $m_j$}
                    $p_i:=\argmin_j\lVert k_i-\phi_j\rVert_2^2$\;
                    $\mathcal{C}_j=\mathcal{C}_j\cup k_{p_i}$\;
                }
                $\phi_j:=\argmin\sum_{k\in\mathcal{C}_j}\lVert k-\phi_j\rVert_2^2$
            }
            Update $\mathcal{M}$\;
            \If{$\mathcal{M}=\mathcal{M_{\text{prev}}}$}{Converged\;}
        }
\end{algorithm}

%------------------------ self attn --------------------------
\begin{figure}[!t]
\centering
\includegraphics[trim={0mm 0mm 0mm 0mm},clip,width=1\linewidth]{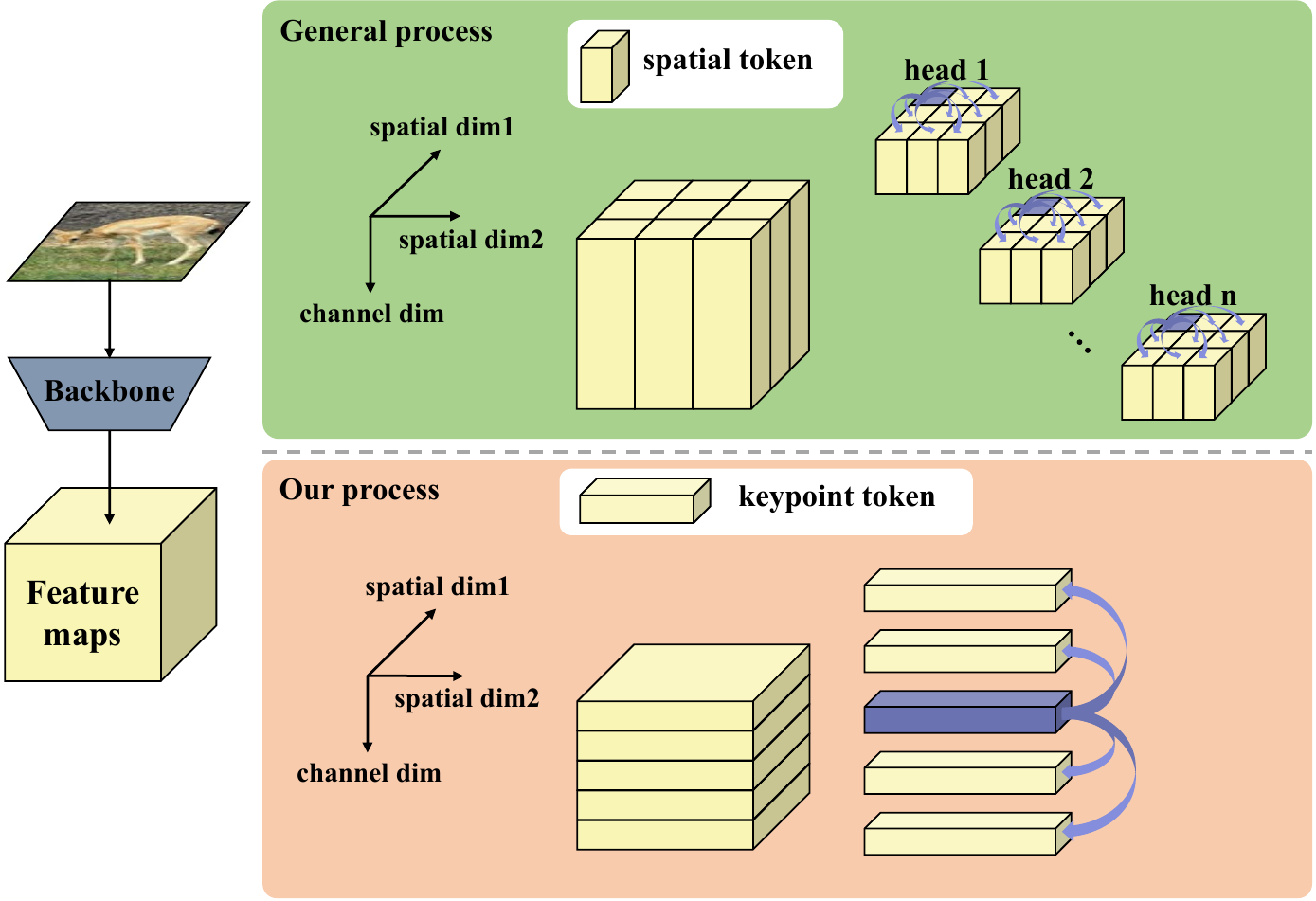}
\caption{
In the vision transformer process, spatial multi-head self-attention works on the split spatial feature patches, where each patch represents a spatial token. Each token is then divided into multiple heads along the channel dimension.
In our design, channel single-head self-attention is explicitly performed in keypoint features. Attention takes the entire feature slice as a token. Relevant keypoints can be involved in the prediction of each keypoint to deliver supporting clues.}
\label{fig:different_self_attn}
\end{figure}

\textbf{NanoBlock.}
Considering the body parts require not only keypoint features but also contextual information, we further propose a lightweight block for modulating image context, called NanoBlock.
In this work, the NanoBlock is built with a two-layer sequential container (Conv-BN-ReLU), with a kernel size equal to 3.
Then, channel alignment is achieved through a linear mapping.
The NanoBlock is simply fed the instance features $F_I\in\mathbb{R}^{N_c\times H\times W}$ produced by the HRNet, where $N_c$ is the number of the output channels.
Therefore, the NanoBlock produce the instance-specific context tokens $T_c$.
After that, the body part biases are fused with $T_c$ to generate instance-level body part prompts $P_{\text{bp}}$.

\subsection{Keypoint-Interactive Encoder}
\label{kit_encoder}
An essential but under-explored problem in pose estimation is how to obtain consistent results both in animal and human pose datasets.
Based on the above considerations, we propose a connecting pattern for Convs and SAs, incorporating the advantages of the two powerful modelling techniques to address the general mammal pose estimation problem.
Our design is different from the structure of canonical ViTs, which uniformly split feature maps into patches, obtaining limited performance on small data volumes.

\textbf{Input tokens.}
The KIT module takes the entire channel-wise feature maps as the keypoint tokens $T_k=\{T_1,T_2,\ldots,T_N\}$ of the Transformer input without any spatial splitting.
Typically, considering the location-sensitive nature of the pose estimation task, 1D positional embedding $E_{\text{pos}}\in\mathbb{R}^{N\times C}$ is added to every specific keypoint token to prompt an attribute order of keypoins in the channel dimension.
% \begin{equation}
% \begin{aligned}
% E_{\text{kpt}}=[T_1,T_2,\ldots,T_N]+E_{\text{pos}}.
% \end{aligned}
% \end{equation}
\begin{equation}
    \begin{aligned}
        E= & [E_{k}\ |\ P_{\text{bp}}] \\
         = & \left[\begin{array}{ccccc|cccc}
                T_1 & T_2 & T_3 & \cdots & T_N & &  &  &  \\
                +   & +   & +   & \cdots & +   &  &      &  & \\
                \undermat{keypoint tokens}{E_{\text{pos}_1} & E_{\text{pos}_2} & E_{\text{pos}_3} & \cdots & E_{\text{pos}_N}} & \undermat{body part prompts}{ P_1 & P_2 & \cdots & P_{N_p} } \\
            \end{array}\right] \\[10pt]
    \end{aligned}
\end{equation}
Together with $P_{\text{bp}}=\{P_1,P_2,\ldots,P_{N_p}\}$, they serve as input tokens for the KIT Encoder.

% --------------------------------------------------
% %Figure :attention maps
\begin{figure}[t] 
\centering
\includegraphics[trim = 0mm 0mm 0mm 0mm, clip, width=\linewidth]{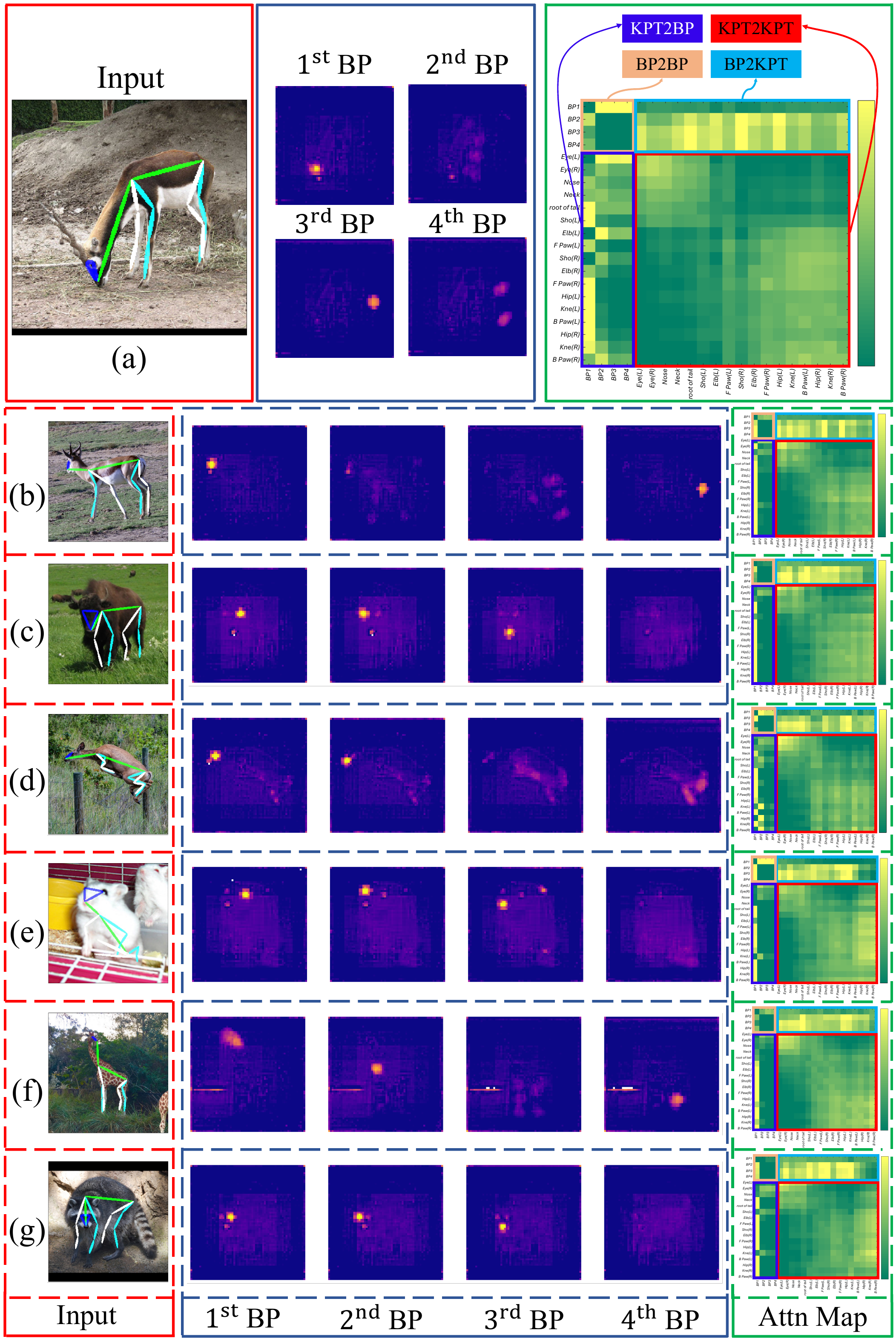}
\caption{Visualisation of the KITPose-E2C4 body part prompts and the interaction patterns. The leftmost column corresponds to the input instances, the middle 4 columns enclosed by blue dashed lines delineate the body part representations, and the rightmost column contains the attention map for information interactions. `BP' and `KPT' mean body part and keypoint respectively. The attention map can be divided into four parts: the orange box signifies BP-to-BP responses, the cyan box represents BP-to-KPT responses, the blue box denotes KPT-to-BP responses, and the red box indicates KPT-to-KPT responses.}
\label{fig:body-parts}
\end{figure}
% --------------------------------------------------

\textbf{Keypoint Interaction of Self Attention.}
SA is the key component of Transformer that captures the global relationships among involved tokens, aggregating the relevant information while suppressing the irrelevant clues.
In our design, we aim to emphasise the interaction among all the keypoints, which are tokenised from the slices of the given feature maps.
Additionally, we involve the body part prompts, which participate in the interaction stage through the self-attention mechanism, providing richer structure-supporting dependencies.
As shown in Fig.~\ref{fig:different_self_attn}, in contrast to the canonical MSAs, we apply single-head self-attention to the Transformer layer to perform interaction among keypoints and body parts.
We introduce our detailed computation procedure as follows.
Considering a sequence of $C$-dimensional input tokens $E=(E_k\ |\ P_{\text{bp}})$.

The input token $E$ is projected to query (Q), key (K), and value (V) as follows,
\begin{equation}
    \begin{aligned}
    q_i=e_iW_q^T,\quad k_i=e_iW_k^T,\quad v_i&=e_iW_v^T,
    \end{aligned}
\end{equation}
where $W_q,W_k,W_v\in\mathbb{R}^{C\times C}$ are three linear projection matrices.
The attention map ($A$) stands for the degree of relevance among involved keypoints, which is computed by a scaled dot-product operation followed by a softmax.
Specifically, each element of the attention map ($A[i,j]$) reflects the relevance between the \emph{i}-th keypoint and the \emph{j}-th keypoint in their representation space.
The attention is a weighted sum of $V$ with the attention map.
Then, the output $O=\{o_1,o_2,\ldots,o_N\}$ is computed by applying the linear projection ($W_o\in\mathbb{R}^{C\times C}$) to the single attention head,
\begin{gather}\label{eqn:attn_map_and_out}
A[i,:]=\text{Softmax}_j(\frac{q_ik_j^T}{\sqrt{C}}).\\
y_i=\sum_j^{K}A[i,j]v_j,\quad o_i=y_iW_o^T
\end{gather}

To intuitively reflect the effect of our design, Fig~\ref{fig:body-parts} shows body part prompts and attention maps of the illustrated samples.
From sample \textit{a} and \textit{b}, we observe that within the same species, pose variations lead to distinct body part prompts.
Consequently, the difference reflects the instance adaptation of our modules via information interactions.
In terms of sample \textit{c} and \textit{e}, we notice that when animals are facing the camera directly, their body part prompts are similar, resulting in a consistent interaction pattern.
While for the extreme poses in sample \textit{d} and \textit{g}, it is evident that the generated body part prompts and interactive patterns vary.
For instance, in sample \textit{f}, when the giraffe's head interacts with other parts, the response values are notably small, primarily due to the considerable distance between the head and other body parts.
Taking a holistic view of the attention maps, interactions between keypoints exhibit a rough diagonal pattern, reflecting relations to adjacent keypoints.
While there are nearly no interactions among different body parts.
Besides, from the horizontal striped pattern, the body parts prompts exhibit strong structural-supporting connections with keypoints.

\subsection{Keypoint-Interactive Transformer Block}
The KIT block consists of a stack of identical layers, where each layer is composed of two sub-layers.
The first sub-layer performs keypoint interaction via self-attention.
The second sub-layer is a canonical Feed-Forward Network (FFN), composed of two linear projections separated by a GeLU activation.
As shown in Fig.~\ref{fig:pipeline},
the first linear projection is conducted to expand the hidden dimension from $d$ to $3d$, and the second linear projection reduces the dimension from $3d$ back to $d$.
Complementary to the SA module, the FFN performs dimensional expansion/reduction and non-linear transformation on each token, thereby enhancing the representation capacity of each token.
% To further smooth the information flow among layers within the Transformer, we introduce \textit{dense} connections to fuse subsequent layers.
% Fig.~\ref{fig:pipeline} illustrates the multiple densely connected KIT blocks.

\subsection{Keypoint weighting strategy}
The widely used loss function for heatmap regression is the Mean Square Error (MSE).
It minimises the pixel-wise divergence between the predicted heatmap and the ground-truth heatmap.
However, it fails to manage the imbalance reflected by different keypoints.
Intuitively, keypoints at different body parts are semantically unique, providing diverse contributions to pose estimation, especially for animal pose datasets.
In principle, existing research has not delved into the effect of these keypoint weights on pose estimation for now.
We, therefore, explore three different keypoint weight strategies,~\ie the hand-crafted, constrained, and adaptive manners.

\textbf{Hand-crafted weight strategy}.
The hand-crafted weight strategy has been widely used in pose estimation, which can be formulated as:
\begin{equation}
    \mathcal{L}_{MSE}=w_i\cdot\sum_{i=1}^N\Vert H^i-\hat{H^i}\Vert_2^2,
\end{equation}
where the factor $w_i$ is the weight of the $i$-th keypoint.
It is a straightforward way to address the imbalance issue by assigning different weights to different keypoints.
However, the hand-designed weight factors are sub-optimal.

\textbf{Constrained weight strategy}.
A feasible way is to impose constraints on pre-defined weights in the MSE loss.
The prior distribution is incorporated to constrain the weights.
We add an $\ell$-2 regulariser here, and the overall regression loss is formulated as follows:
\begin{equation}
\mathcal{L}=\sum_i^Nw_i\Vert H^i-\hat{H}^i\Vert_2^2+\lambda\sum_i^N\Vert w_i-1\Vert_2^2,
\end{equation}
where $\lambda$ is a balancing parameter ($\lambda=0.01$ in our implementation).
Although the imbalance is mitigated to some extent by introducing the constrained keypoint weights, existing animal datasets generally consist of individual images that are not related to each other.
In other words, these images usually contain a large variety of poses with inconsistent numbers of well-detected and hard keypoints.

\textbf{Adaptive weight strategy}.
A step further, \cite{lin2017focal} introduces the focal loss to address the imbalance issue for binary classification. % \att{XXXcite}
% This design can be easily extended to a more complex task, such as pose estimation.
To be consistent with the heatmap regression formulation, we convert the modulating factor over the discrete representation into a continuous domain, by representing the scaling factor $(1-p_t)^\gamma$ into an error term \iffalse$|H^i-\hat{H^i}|^\gamma$\fi, the $i$-th estimated keypoint regression weight map $W^i$ can be presented as:
% We propose a new adaptive weighting map $W$, in which the loss weights are written as a focal style:
\begin{equation}
W^i=\lvert H^i-\hat{H^i}\rvert^\gamma=\mathit{e}^\gamma,
\end{equation}
where $\gamma\ge 0$ is the hyper-parameter for adjusting different keypoints.
When the pose estimation of a keypoint is inaccurate, \ie deviated away from the true position $y$, the absolute error $\mathit{e}$ is relatively large, thus it pays more attention to learning this hard keypoint.
Otherwise, the error term goes to 0 and the loss for well-detected keypoint is down-weighted.
The focusing parameter $\gamma$ controls the down-weighting rate smoothly (in practice, we set $\gamma=2$).
Fig.~\ref{fig:weight_map} shows that the adaptive weight strategy can further introduce instance- and keypoint-specific importance to our model.
We believe that the corresponding weight strategy is valuable in demonstrating its potential for general animal pose estimation.

\section{Experiments}
\subsection{Datasets and evaluation metrics}
We evaluate the performance of our KITPose on \textbf{AP10K}~\cite{yu2021ap} and \textbf{AnimalPose}~\cite{cao2019cross}.
The former is the first large-scale benchmark for animal pose estimation, containing 10k images collected and filtered from 23 animal categories and 54 species with 17 keypoints, while the latter is the common benchmark containing over 4k images from 5 species with 20 keypoints.
We report the numerical results on the corresponding \textit{val} sets for ablation studies and comparisons with other state-of-the-art methods.
In the following evaluation, we employ the mean Average Precision (AP) defined based on the Object Keypoint Similarity (OKS) as the primary evaluation metric on the AP10K~\cite{yu2021ap} and AnimalPose dataset.
Average Recall and APs at different OKS thresholds are also reported, \ie AR, AP.5 and AP.75, as well as APs for different instance sizes, \ie AP$^{M}$ and AP$^{L}$.

We also conduct experiments on \textbf{AnimalKingdom}~\cite{ng2022animal} dataset with Percentage of Correct Keypoint (PCK) as the evaluation metric.
The AnimalKingdom pose estimation dataset contains a common set of keypoints across five classes: mammals, reptiles, amphibians, birds, and fishes, obtained from a wide range of real-world activities.
There are over 7k mammal samples with 23 keypoints in the AnimalKingdom dataset, which differs from AP10K.
After analysing the dataset, we find that there are a total of 145 classes of mammals. 
This indicates that this dataset contains fewer mammal samples per category but a greater diversity of species, making it more challenging.
In addition, we follow the detailed protocol defined in~\cite{ng2022animal}.
PCK@$\alpha$ provides a quantitative measure for localisation accuracy of predicted keypoints by allowing the acceptable error between the predicted keypoints and their ground truth locations, which can be adjusted to reflect the scale and variability of the instances.
In the evaluation stage, the PCK@$0.05$ ($\alpha=0.05$) score is reported, which computes the percentage of correct keypoints whose distance to ground truth is within $\alpha$ times the longest bounding box side.

\subsection{Model settings}
In general, ImageNet~\cite{deng2009imagenet} can provide general visual knowledge and is widely used for backbone pretraining.
Unless specified, Imagenet pretrained HRNet~\cite{sun2019deep} is used as the initialised backbone in our implementations.
Specifically, HRNet-W32 and HRNet-W48 are employed to evaluate the model capacity of different model sizes, reflecting the parameter volume-relevant performance changes.
% The weights pre-trained on ImageNet~\cite{deng2009imagenet} is used to initialise the backbone.
The KIT encoder is initialised with random parameters in truncated normal distribution instead.
All the encoder embedding size is set to 1024 based on a trade-off between the model complexity and accuracy.
Drawing on this, we name the models with abbreviations for convenience.
In particular, KITPose-E$\#i$C$\#j$ means the number of encoder layers is $\#i$, and the number of the clusters is $\#j$.

\subsection{Implementation Details}
In our implementation, we follow the top-down pose estimation paradigm.
We use the ground-truth bounding box annotations in AP10K to crop images with single animal instances to a fixed size.
% We adopt ImageNet and COCO as general and human-prior pre-training datasets, respectively.
% For the keypoint features extractor in the pose estimation pipeline, we select the mainstream HRNet and initialise it with the pre-training weights.
The keypoints are predicted from the heatmaps by forward-passing the backbone and the Keypoint-Interactive Transformer.
To obtain pixel-level keypoints' locations, the coordinate decoding approach~\cite{zhang2020distribution} is used to alleviate the quantisation error.
% Our KIT benefits from handling multiple species in animal pose estimation, enhancing the model capacity.
We train and test the model on a single NVIDIA RTX $3090$Ti GPU.

\textbf{Hyper-parameters setting.} During training, we follow the default settings as in HRNet~\cite{sun2019deep}, \ie we configure $256\times256$ and $384\times384$ as two input resolution choices, for a fair comparison with existing methods.
We train the entire model using an adaptive MSE loss and an ADAM~\cite{kingma2014adam} optimiser.
We set the initial learning rate to $5$e-$4$ for the backbone and KIT module, with a weight decay of $1$e-$4$.
$\beta1$ and $\beta2$ are set to $0.9$ and $0.999$ by default.
We also use a multi-step decay scheduler, dropping the learning rate to $5$e-$5$ and $5$e-$6$ at the $200$th and $230$th epochs, respectively, end at the $250$th epoch.

\textbf{Data augmentation.}
Our data augmentation settings are consistent with existing design~\cite{mmpose2020}.
Simple augmentations include random rotation ($-40^\circ\sim40^\circ$), random scale (0.5$\sim$1.5), and horizontal flipping test.
Advanced augmentations include half-body transformation~\cite{sun2019deep} and cutmix~\cite{yun2019cutmix}.
Cutmix forces the model to focus on less discriminative parts of the instance, aligning with the concept of structure-supporting learning by the KIT module.

% --------------------------------------AP10K Performance--------------------------------------
\begin{table*}[t]
\renewcommand{\arraystretch}{1.2}
\centering
\begin{tabular}{lccccccccccc}
\toprule
Methods   & Source  & Backbone   & Input size     & \#Params & GFLOPs & AP   & AP$^{.50}$ & AP$^{.75}$ & AP$^{M}$ & AP$^L$ & AR \\
\midrule\midrule 
       % Sim.Base.~\cite{xiao2018simple}    & ECCV2018  & ResNet50  &  $256\times256$    & 34.0M & 11.99 & 56.8  & 88.0  & 56.0  & 47.3  & 57.1  & 62.0 \\
       Sim.Base.~\cite{xiao2018simple}    & ECCV2018  & ResNet101  &  $256\times256$    & 53.0M & 16.50 & 70.6  & 94.1  & 76.6  & 54.4  & 71.0  & 74.0 \\
       Sim.Base.~\cite{xiao2018simple}    & ECCV2018  & ResNet101  &  $384\times384$    & 53.0M & 37.13 & 71.7  & 94.7  & 77.4  & 47.2  & 72.3  & 75.1 \\
       HRNet~\cite{sun2019deep} & CVPR2019      & HRNet-W32      & $256\times256$ & 28.5M   & 9.49   & 73.8 & 95.8      & 80.3      & 52.3     & 74.2   & 76.9 \\
       % HRNet~\cite{sun2019deep}        & HRNet-W32    & COCO2017   & $256\times256$ & 28.5M   & 9.49   & 75.3 & 96.2      & 82.7      & 61.6     & 75.6   & 78.8 \\
       HRNet~\cite{sun2019deep}  & CVPR2019      & HRNet-W32       & $384\times384$ & 28.5M   & 21.36  & 74.7 & 95.8      & 81.8      & 54.4     & 75.2   & 78.4 \\
       HRNet~\cite{sun2019deep}   & CVPR2019     & HRNet-W48     & $256\times256$ & 63.6M   & 19.49  & 74.9 & 96.2      & 81.7      & 56.1     & 75.2   & 78.6 \\
       HRNet~\cite{sun2019deep}   & CVPR2019     & HRNet-W48      & $384\times384$ & 63.6M   & 43.84  & 75.6 & 96.5      & 82.6      & 58.7     & 76.0   & 79.3 \\
       DARK~\cite{zhang2020distribution}  & CVPR2020      & HRNet-W32     & $256\times256$ & 28.5M   & 9.49   & 74.6 & 95.9      & 80.8      & 62.6     & 74.9   & 77.8 \\
       DARK~\cite{zhang2020distribution}    & CVPR2020  & HRNet-W32 &   $384\times384$  & 28.5M    & 21.36 & 76.3   & 96.2  & 83.0  & 57.9  & 76.8  & 79.3 \\
       TokenPose~\cite{li2021tokenpose} & ICCV2021  & B           & $256\times256$ & 19.1M   & 6.65   & 69.2 & 93.5      & 75.9      & 54.7     & 69.7   & 72.7 \\
       TokenPose~\cite{li2021tokenpose} & ICCV2021   & L/D24       & $256\times256$ & 35.8M    & 14.64  & 72.7 & 94.3      & 79.2      & 60.6     & 72.7   & 75.7 \\
       % TokenPose~\cite{li2021tokenpose}   & L/D24       & COCO2017   & $256\times256$ & 35.8M    & 14.64  & 75.1 & 95.8      & 81.6      & 59.7     & 75.5   & 78.2 \\
       TransPose~\cite{yang2021transpose} & ICCV2021   & H-A6        & $256\times256$ & 16.9M    & 11.25  & 74.2 & 95.7      & 80.8      & 56.2     & 74.6   & 77.0 \\
       HRFormer~\cite{yuan2021hrformer} & NIPS2021   & HRFormer-B    & $256\times256$ & 43.2M    & 18.35  & 74.5 & 95.9      & 81.6      & 55.8     & 74.9   & 77.6 \\
    %   HRFormer    & HRFormer-B  & COCO2017   & $256\times256$
       HRFormer~\cite{yuan2021hrformer} & NIPS2021   & HRFormer-B    & $384\times384$ & 43.2M    & 37.82  & 75.7 & 96.4      & 82.3      & 54.7     & 76.2   & 78.9 \\
       ViTPose-B~\cite{xu2022vitpose} & NIPS2022      & ViT-B        & $256\times192$ & 89.9M   & 17.85  & 66.3 & 92.0      & 72.5      & 59.4     & 66.4   & 70.0 \\
       ViTPose-L~\cite{xu2022vitpose}  & NIPS2022     & ViT-L         & $256\times192$ & 308.5M  & 59.78  & 71.2 & 94.2      & 77.8      & 59.5     & 71.4   & 74.6 \\
       CLAMP~\cite{zhang2023clamp} & CVPR2023  & ViT-B  & $256\times256$ & - & - & 74.3 & 95.8 & 81.4   & 47.6  & 74.9  & 77.5 \\
\midrule KITPose-E2C4 & Ours  & HRNet-W32   & $256\times256$ & 68.4M    & 9.51   & 76.6 & 96.7      & 84.5     & 57.5    & 76.9   & 79.5 \\
       % KITPose-E2  & HRNet-W32    & COCO2017   & $256\times256$ & 57.9M    & 9.50   & 78.9 & 97.1      & 86.3      & 60.4     & 79.1   & 82.0 \\
       KITPose-E2C4 & Ours & HRNet-W48     & $256\times256$ & 103.5M   & 19.50  & 77.3 & 96.3 & 84.9 & 57.0 & 77.7   & 80.4 \\
       % KITPose-E2  & HRNet-W32    & COCO2017   & $384\times384$ & 68.4M   & 21.37  & 79.9 & \textbf{97.7}      & 86.8    & 58.9     & 80.2   & 82.7 \\
       KITPose-E2C4 & Ours & HRNet-W32      & $384\times384$ & 115.6M    & 21.38  & 77.2 & \textbf{97.3}      & 85.0      & \textbf{56.4}    & 77.6   & 80.3 \\
       KITPose-E2C4 & Ours  & HRNet-W48 & $384\times384$    & 237.2M    & 43.88 & \textbf{77.9}  & 97.1  &\textbf{85.4}   &54.4   &\textbf{78.3}   &\textbf{80.7} \\   
       \midrule
       KITPose-E2$\dag$ & Ours  & HRNet-W32   & $256\times256$ & 71.5M    & 9.52   & 75.8 & 96.3      & 83.3     & 54.8    & 76.1   & 78.8 \\
       KITPose-E2$\dag$ & Ours  & HRNet-W48   & $256\times256$ & 106.6M    & 19.52   & 76.5 & 96.7      & 83.5     & 54.0    & 76.8   & 79.5 \\
\bottomrule
\end{tabular}
\vspace{1em}
\caption{Performance comparison on AP10K validation set. In particular, the improvement of our approach for AP$^{.75}$ is promising compared to other approaches. Params and GFLOPs are calculated to measure the model volume and complexity. $\dag$ denotes that training on a mixed animal dataset (AP10K, Animal Kingdom, AnimalPose).}
% \vspace{-1em}
\label{tab:ap10k} 
\end{table*}
% -----------------------------------------------------------------------------------------------

% ------------------------------------AnimalPose Performance------------------------------------
\begin{table}[htbp]
\renewcommand{\arraystretch}{1.2}
\setlength\tabcolsep{2pt} % set the column space
\centering
\begin{tabular}{llcccccc}
\toprule
Methods  & Backbone & AP   & AP$^{.50}$ & AP$^{.75}$ & AP$^{M}$ & AP$^L$ & AR \\
\midrule\midrule 
% Sim.Base.~\cite{xiao2018simple} & ResNet50 & 68.7  & 93.7  & 76.9  & 63.7  & 69.9  & 73.0  \\
Sim.Base.\cite{xiao2018simple}  & ResNet101  & 69.6  & 94.8  & 77.4  & 66.7 & 70.5 & 73.6  \\
HRNet~\cite{sun2019deep}    & HRNet-W32    & 74.0  & 95.9  & 83.3  & 69.6  & 75.1  & 78.0  \\
HRNet~\cite{sun2019deep}    & HRNet-W48    & 74.2  & 94.8  & 83.3  & 70.1  & 75.6  & 78.1  \\
DARK~\cite{zhang2020distribution}  & HRNet-W32 & 74.6  & 95.7  & 83.2  & 71.3  & 75.5  & 78.4  \\
DARK~\cite{zhang2020distribution}  & HRNet-W48 & 74.4  & 95.8  & 83.4  & 70.0  & 75.6  & 78.2  \\
% TokenPose~\cite{li2021tokenpose}    & B    & 70.0  & 92.7  & 78.9  & 67.6  & 70.5  \\
TokenPose~\cite{li2021tokenpose}    & L/D24 & 71.8  & 94.8  & 81.5  & 69.0  & 72.5  & 76.1  \\
HRFormer~\cite{yuan2021hrformer}    & HRFormer-B    & 73.4  & 94.7  & 82.8  & 70.2  & 74.4  & 77.0  \\
ViTPose~\cite{xu2022vitpose}    & ViT-B & 67.4  & 92.6  & 74.3  & 66.3  & 68.0  & 71.6  \\
ViTPose\cite{xu2022vitpose}    & ViT-L & 66.2  & 91.5  & 73.0  & 66.7  & 66.4  & 70.4  \\
CLAMP~\cite{zhang2023clamp}    & ResNet50  & 72.5  & 94.8  & 81.7  & 67.9  & 73.8  & 76.7  \\
CLAMP~\cite{zhang2023clamp}    & ViT-B & 74.3  & 95.8  & 83.4  & \textbf{71.9}  & 75.2  & 78.3  \\
\midrule KITPose-E2C5   & HRNet-W32   & 75.1    & 95.8  & 83.8  & 69.6  & 76.7  & 79.2  \\
KITPose-E2C5    & HRNet-W48 & \textbf{76.7}  & \textbf{96.8}  & \textbf{86.8}  & 71.7  & \textbf{77.9}  & \textbf{80.3}  \\
\bottomrule
\end{tabular}
% \vspace{1em}
\caption{Performance comparison on AnimalPose validation set. The input resolution of all methods is 256 × 256.}
\label{tab:animalpose} 
\vspace{-2em}
\end{table}
% -----------------------------------------------------------------------------------------------

% ---------------------------------- Animal Kingdom Performance ------------------------------------
\begin{table}[!t]
\renewcommand{\arraystretch}{1.2} % set the space between rows
\centering
\setlength\tabcolsep{1pt} % set the column space
\footnotesize
\begin{tabular}{lcccccccccc}
\toprule
Methods     & Hea & Sho & Elb & Wri & Hip & Kne & Ank & Mou & Tail & Mean \\ \midrule\midrule
ResNet-50~\cite{xiao2018simple} & 60.0 & 49.7   & 43.2  & 49.0  & 48.8  & 43.7  & 49.8  & 68.7  & 46.4  & 54.7 \\
ResNet-101~\cite{xiao2018simple} & 59.3 & 51.7  & 45.0  & 51.8  & 45.6  & 43.2  & 48.8  & 71.1  & 48.0  & 55.6 \\
HRNet-W32~\cite{sun2019deep}    & 58.5 & 50.6   & 47.3  & 50.9  & 49.1  & 47.7  & 46.6  & 71.2  & 48.9  & 57.1 \\
% HRNet-W32*~\cite{sun2019deep}    & 60.6 & 51.8   & 54.0  & 63.8  & 54.1  & 52.6  & 56.5  & 75.6  & 52.9  & 62.2 \\
HRNet-W48~\cite{sun2019deep}    & 61.0 & 49.6   & 46.6  & 51.3  & 48.9  & 43.9  & 45.6  & 73.6  & 49.8  & 57.3 \\
DARK-HR32~\cite{zhang2020distribution} & 61.1  & 50.1  & 47.7  & 54.7  & 49.7  & 46.9  & 49.8  & 71.3  & 51.2  & 58.0 \\ 
TokenPose~\cite{li2021tokenpose}   & 60.1 & 48.6   & 48.2  & 52.7  & 47.4  & 45.6  & 49.2  & 69.6  & 49.6  & 56.5 \\
TransPose~\cite{yang2021transpose}   & 59.8 & 47.0   & 46.3  & 54.6  & 51.5  & 46.0  & 46.5  & 73.7  & 49.7  & 57.6 \\
HRFormer-S~\cite{yuan2021hrformer}    & 61.9 & 50.5   & 46.7  & 55.2  & 43.6  & 46.5  & 51.7 & 72.9  & 51.0  & 57.4 \\
HRFormer-B~\cite{yuan2021hrformer}    & 62.0 & 51.6   & 49.5  & 57.2  & 49.5  & 46.3  & 49.8  & 74.7  & 51.2  & 58.7 \\
ViTPose-B~\cite{xu2022vitpose} & 60.6  & 53.0  & 48.5  & 56.0  & 51.4  & 46.0  & 51.1  & 68.6  & 49.1  & 57.1  \\
ViTPose-L~\cite{xu2022vitpose} & 57.7  & 48.9  & 46.6  & 52.7  & 47.1  & 45.0  & 49.0  & 66.4  & 46.5  & 54.5  \\
CLAMP-R50~\cite{zhang2023clamp}    & 58.7  & 52.3  & 45.0  & 53.0  & 49.1   & 45.5 & 49.8 & 69.8    & 48.9   & 56.0 \\
CLAMP-ViTB~\cite{zhang2023clamp}   & 63.1  & 53.5  & 51.3  & 58.0  & 51.0  & 46.8  & 51.1  & 70.8  & 53.3  & 58.9  \\
% CLAMP-ViTL~\cite{zhang2023clamp}   & 61.3  & 51.2  & 52.0  & 60.8  & 52.1  & 50.0  & 56.4  & 74.9  & 53.9  & \textbf{60.8}  \\
\midrule
KITPose-E2C6    & 60.3  & 50.1  & 48.0  & 54.7  & 48.9  & 47.6  & 51.2  & 73.5  & 51.6  & 58.8 \\
KITPose*-E2C6  & 63.8  & 49.2  & 49.8  & 56.0  & 51.1  & 48.4  & 54.7  & 75.3  & 50.8  & \textbf{59.1} \\
% KITPose-E2* & \textbf{64.9}  & \textbf{52.5}  & \textbf{54.4}  & \textbf{65.4}  & \textbf{58.0}  & \textbf{53.3}  & \textbf{57.4}  & \textbf{75.7}  & \textbf{54.1}  & \textbf{63.5} \\
\bottomrule
\end{tabular}%
% \vspace{1em}
\caption{Performance comparisons on the AnimalKingdom mammals test set (``Mean" is PCK@$0.05$). The input resolution of all methods is $256\times256$. KITPose and KITPose* are equipped with the HRNet-W32 and HRNet-W48 backbone, respectively.}
\label{tab:animalkingdom}
\vspace{-2em}
\end{table}
% -----------------------------------------------------------------------------------------------

\subsection{Results and Analysis}
\textbf{Evaluation on AP10K Dataset.}
Table~\ref{tab:ap10k} reports the performance comparisons among our design and other state-of-the-art methods, involving ConvNets, Transformers, and hybrid structures.
On the AP10K \textit{validation} set, as we can see, in comparison with the state-of-the-art Conv-based architectures (\ie SimpleBaseline, HRNet and DarkPose), KITPose outperforms the representative solutions with comparable or fewer computation complexity, surpassing $6.0$ AP than SimpleBaseline (KITPose-E2C4 $76.6$ v.s. Res-101 $70.6$), $2.8$ AP than HRNet (KITPose-E2C4 $76.6$ vs. HRNet-W32 $73.8$) and $2.0$ AP than DarkPose (KITPose-E2C4 $76.6$ vs. DarkPose-W32 $74.6$).
% Looking into the series of Transformer-based methods from Table~\ref{tab:hp_method} and Table~\ref{tab:ap10k}, \att{we can see that previous pose estimation approaches ignore the diversity of animal species, impeding their generalisation ability, and thus are not suitable for general mammal pose estimation tasks.}
Our Transformer-based KITPose achieves the new state-of-the-art performance.
Compared with the hybrid architectures (\ie TokenPose and TransPose), KITPose achieves superior performance with lower computational complexity, despite its larger model size.
With fewer computation budget, KITPose-E2C4 achieves $3.9$ and $2.4$ AP improvement over TokenPose and TransPose, respectively (KITPose-E2C4 $76.6$ vs. TokenPose-L/D24 $72.7$ and TransPose-H-A6 $74.2$).
% in comparison to the TokenPose-L/D24~\cite{li2021tokenpose} and TransPose-H-A6~\cite{yang2021transpose} that use HRNet-W48 as the backbone, HRNet-W48 achieves a higher $74.9$ AP.
Compared with the plain Transformer-based architectures (\ie HRFormer and ViTPose), KITPose achieves better pose estimation performance against HRFormer (KITPose-E2C4 $76.6$ vs. HRFormer-B $74.5$), while using lower computational complexity and requiring less training time.
Since the original ViTPose~\cite{xu2022vitpose,xu2022vitpose+} employs the multi-task training scheme, for the fair comparison, we follow their implementations and conduct the training process on the AP10K dataset to obtain the final results.
Therefore, compared with the human pose estimation SOTA method ViTPose-L, the proposed KITPose-E2C4 obtains significant improvements,~\ie $5.4$ AP gain with a considerable reduction in model parameters and GFLOPs.

% Apart from the above practice, we make comparisons with the pure Transformer architecture, HRFormer and ViTPose.
% The proposed KITPose-E2 with HRNet-W32 as backbone outperforms HRFormer-B by $+1.6$ AP with less computation cost.
% Compared with the human poes estimation SOTA method ViTPose-L, the proposed KITPose model using HRNet-W32 as backbone obtains significant improvements,~\ie $4.9$ points gain with a considerable reduction in model parameters and GFLOPs.
% It should be noted that all the Transformer blocks in our KITPose are trained from scratch, without any pre-trained parameters given.
% We argue that pure transformer-based models do not achieve comparable results or even get worse in the animal domain, when facing the small data volume and large data variances.

In addition, considering the large resolution setting ($384\times384$), our KITPose-E2C4 based on HRNet-W32 and HRNet-W48 achieves $77.2$ and $77.9$ AP scores, which exhibit $+0.6$ improvement compared to those of input size $256\times256$.
Besides, we find that our KITPose outperforms at least $2.0$ AP points against the existing state-of-the-art methods.
It should be noted that our design with $256\times256$ input also surpasses the performance of existing methods at an input size of $384\times384$.
For example, KITPose-E2C4 at $256\times256$ obtains $+0.9$ AP than HRFormer-B at $384\times384$.
% In comparison to DarkPose~\cite{zhang2020distribution} which uses HRNet-W32 as the backbone, our KITPose-E2 attains $1.2$ points gain in terms of AP.
% Using a more powerful Transformer module
% Compared with the representative top-down methods reported on AP10K, the KITPose-E4 achieves the best performance of 77.9 AP points.
It demonstrates that at lower input resolution, the KIT module can provide structural-supporting dependencies to compensate for the lack of pixel and appearance cues, thus surpassing the higher-resolution counterparts.
Furthermore, we compare KITPose with the CLAMP, a cross-modal architecture that incorporates text prompts.
While CLAMP achieves remarkable results under the few/zero-shot paradigm, our method outperforms $2.3$ AP score under the supervised learning paradigm.
Therefore, the necessity and advantages of performing structure-supporting dependencies learning can be obtained.

To evaluate the performance of KITPose on general mammal pose estimation comprehensively, we also report the performance of KITPose-E2$\dag$ on the AP10K val set.
Note that the KITPose-E2$\dag$ model is trained on the mixing of 3 different animal datasets (AP10K, Animal Kingdom, AnimalPose) and directly evaluated on the corresponding dataset, without re-training on them.
It demonstrates the generalisation capability of our method.
The number of keypoints in the model is set to the maximum value among the mixed datasets.
We then follow the FFN splitting strategy in ViTPose+, and adopt the same idea to split FFN in the KIT module into task-agnostic and task-specific parts to encode common and specific knowledge for general mammal pose estimation.
Additionally, an attention mask is incorporated into the KIT module to against the disturbances of redundant tokens.
As shown in the last two rows of Table~\ref{tab:ap10k}, our KITPose-E2$\dag$ model obtains competitive performance at 75.8 AP.
Despite not being as effective as training solely on AP10K when the input size is $256\times 256$, it significantly outperforms the existing state-of-the-art method DARK by 1.2 AP.

\begin{table}[t]
\renewcommand{\arraystretch}{1.2}
\setlength\tabcolsep{4.5pt} % set the column space
\centering
\begin{tabular}{c|c|c|lcccc}
\toprule
A.W.    & KIT   & BP    & AP   & AP$^{.50}$ & AP$^{.75}$ & AP$^{M}$ & AP$^L$ \\
\midrule\midrule 
\xmark  & \xmark  & \xmark  & 73.8  & 95.8  & 80.3  & 52.3  & 74.2  \\
\cmark  & \xmark  & \xmark  & 74.5(\att{$\uparrow0.7$})  & 95.5  & 81.4  & 58.5  & 74.9  \\
\xmark  & \cmark  & \xmark  & 74.3(\att{$\uparrow0.5$})  & 95.6  & 80.9  & 55.9  & 74.7  \\
\cmark  & \cmark    & \xmark    & 76.1(\att{$\uparrow2.3$})  & 96.6  & 82.8  & 61.4  & 76.5 \\
\xmark  & \cmark    & \cmark    & 75.8(\att{$\uparrow2.0$})  & 96.3  & 82.0  & 57.0  & 76.2  \\
\cmark  & \cmark    & \cmark    & 76.6(\att{$\uparrow2.8$})  & 96.7  & 84.5  & 57.5  & 76.9  \\
\bottomrule
\end{tabular}
% \vspace{1em}
\caption{Ablation studies on components of KITPose, including the adaptive weighting strategy (A.W.), Keypoint Interactive Transformer (KIT), and the body part prompt (BP) respectively. All the results are reported on AP10K dataset.}
\vspace{-1em}
\label{tab:ablation_study} 
\end{table}

%---------- Ablation study of clustering numbers ----------------------
\begin{figure}[!t]
    \centering
    \begin{subfigure}{0.45\linewidth}
        \includegraphics[width=\linewidth]{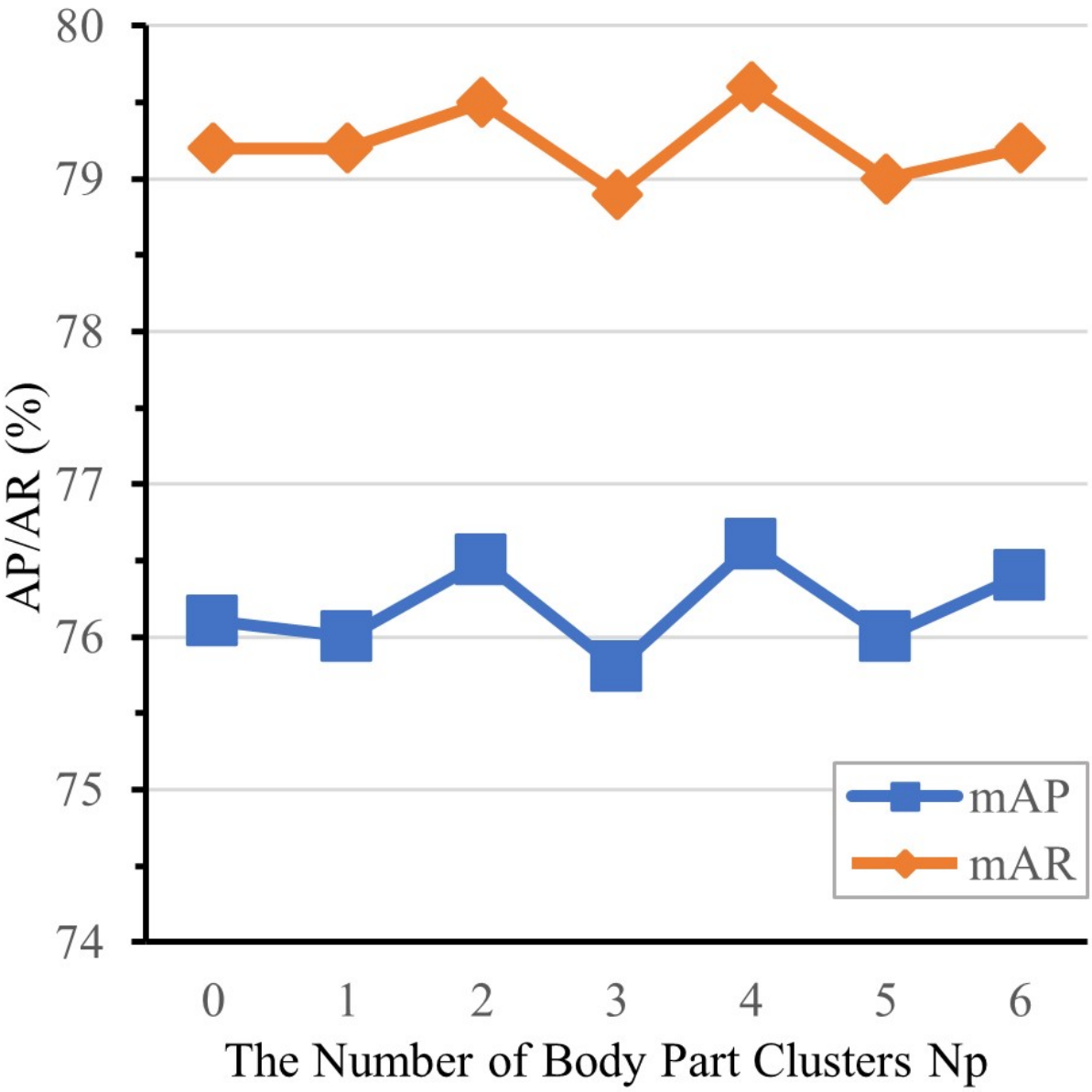} 
        \caption{mAP/AR on AP10K}
    \label{fig:cluster-sub1}
    \end{subfigure}%
    \hspace{2.5em}
    \begin{subfigure}{0.45\linewidth}
        \includegraphics[width=\linewidth]{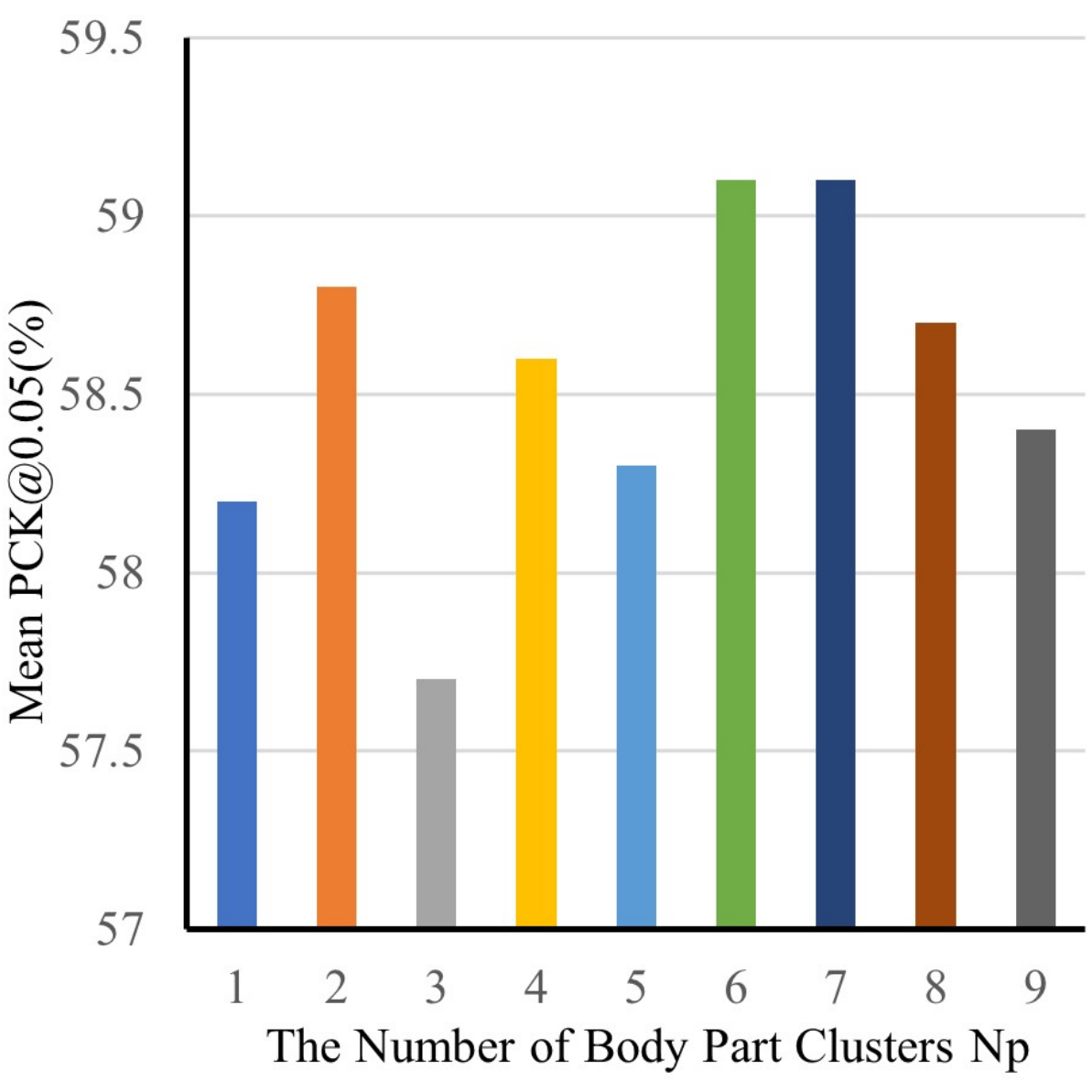}
        \caption{Mean PCK on AK}
        \label{fig:cluster-sub2}
    \end{subfigure}
% \vspace{-10mm}
\caption{Performance sensitivity analysis in terms of the number of body part clusters $N_{p}$. The results are reported on AP10K and AnimalKingdom using the KITPose-E2 model, with an input size of $256\times256$.}
\label{fig:cluster-number}
\vspace{-1em}
\end{figure}
% ---------------------------------------------------------

%------------------------------ Ablation study of body part prompt ----------------------------------
\begin{table}[t]
\renewcommand{\arraystretch}{1.2} % set the space between rows
\centering
\setlength\tabcolsep{1.2pt} % set the column space
\resizebox{0.48\textwidth}{!}{
    \begin{tabular}{c|c|@{\hspace{6pt}}cccccc}
    \toprule
    Hyper-params    & Value &   AP  & AP$^{.50}$    & AP$^{.75}$    & AP$^{M}$  & AP$^{L}$   & Speed (fps) \\ \midrule\midrule
    \multirow{5}{*}{\shortstack{Prompt Number $N_p$ \\ ($\mathcal{D}$=Euclidean, $\tau$=1e-6)}}
        & 1 & 76.0 & 96.2 & 83.2 & 62.7 & 76.3   & 548.0 \\
        & 2 & 76.5 & 96.7 & 84.7 & 60.0 & 76.9   & 482.9 \\
        & 3 & 75.8 & 96.1 & 82.3 & 57.3 & 76.1   & 442.0 \\
        & 4 & 76.6 & 96.7 & 84.5 & 57.5 & 76.9   & 441.5 \\
        & 6 & 76.4 & 96.6 & 83.2 & 60.7 & 76.7   & 399.6 \\
        \midrule
    \multirow{2}{*}{\shortstack{Distance Metric $\mathcal{D}$ \\ ($N_p$=4, $\tau$=1e-6)}}
    % \hspace{0.6em}
        & Cosine & 75.8  & 96.1  & 83.2  & 58.4  & 76.1 & 456.7 \\
        & Euclidean  & 76.6  & 96.7  & 84.5  & 57.5  & 76.9 & 441.5 \\ \midrule
    \multirow{3}{*}{\shortstack{Threshold $\tau$ \\ ($N_p$=4, $\mathcal{D}$=Euclidean)}}
        & 1e-2 & 75.0 & 96.4 & 80.6 & 55.1 & 75.4   & 437.0 \\
        & 1e-4 & 76.2 & 97.0 & 83.4 & 57.5 & 76.6   & 454.5 \\
        % & 1e-6 & 76.6 & 96.7 & 84.5 & 57.5 & 76.9   & 441.5 \\
        & 1e-8 & 76.6 & 96.8 & 84.8 & 61.0 & 76.9   & 433.6 \\
    \bottomrule
    \end{tabular}
}
% \vspace{1em}
\caption{Ablation studies of body part prompt. Results are reported on KITPose-E2 with the input size of $256\times 256$ on the AP10K val dataset. `Distance Metric' refers to the measurement between datapoints during clustering. `Threshold' signifies a convergence criterion where if the cluster medoids shift is less than the value. Speed stands for frame per second during evaluation on a single 3090Ti GPU.}
\label{tab:bp-params}
\vspace{-2em}
\end{table}
%----------------------------------------------------------------------------------------------------------
%------------------------------ Ablation study of encoder layers -------------------------------
\begin{figure}[!t]
\centering
\includegraphics[width=\linewidth]{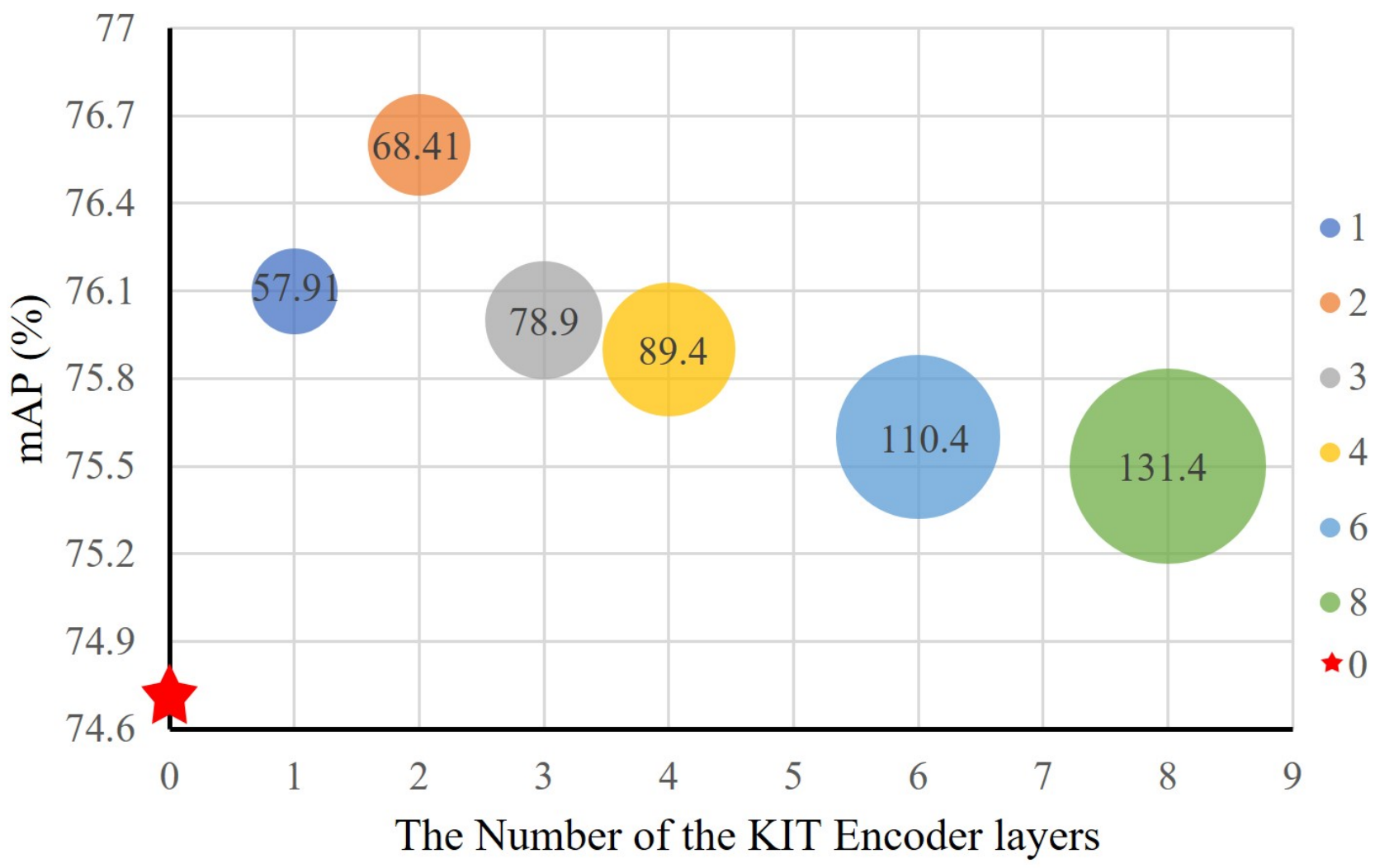}
\caption{Comparisons of KITPose-C4 with different encoder layers on AP10K val set. All backbones are trained with HRNet-W32. The size of each bubble represents the number of model parameters (Millions). The asterisk symbol indicates the baseline without using the KIT module.}
\label{fig:encoder-layers}
\vspace{-1em}
\end{figure}
%-------------------------------------------------------------------------------------------------
\textbf{Evaluation on AnimalPose Dataset.}
In this part, we compare the results of our KITPose with other state-of-the-art methods on the AnimalPose dataset.
The input resolution of all models is set to $256\times256$.
The results for CLAMP are directly quoted from the original paper, while others are the results obtained by training on AnimalPose with their official codes.
Considering that the number of predefined keypoints in AnimalPose is more than that in AP10K, we choose 5 body parts for the experiments, \ie C5.
From Table~\ref{tab:animalpose}, the KITPose-E2C5 reaches 76.7 AP, which is higher than existing state-of-the-art results.
For instance, we outperform the baseline HRNet-W48 by 2.5 AP and the state-of-the-art DARK by 2.1 AP.
We notice that the existing methods with more complex backbones, such as DARK-W48 and ViTPose-L, perform worse than DARK-W32 and ViTPose-B with the AP dropping by 0.2 and 1.2, respectively.
It also demonstrates that KITPose is effective and data-efficient in handling the cross-species problem for improving general mammal pose estimation performance.

\textbf{Evaluation on AnimalKingdom Dataset.}
To test KITPose in more challenging scenarios, we compare it with recent works on the AnimalKingdom dataset.
% In this part, we present the performance comparisons on the AnimalKingdom dataset.
The input resolution of all models is set to $256\times256$.
We follow the training and testing process which is the same as that on the AP10K dataset.
In particular, we train HRNet according to the official code, which achieves $58.0$ and $57.8$ PCK@$0.05$ scores, respectively, 
% Although the results differ from those presented in the original paper~\cite{ng2022animal} (HRNet-W32 $61.59$), they are 
consistent with the reported results by MMPose~\cite{mmpose2020}.
Therefore, we adopt these results for comparison.
In Table~\ref{tab:animalkingdom}, KITPose-E2C6 equipped with HRNet-W32 achieves a $58.8$ PCK@$0.05$ score, outperforming HRNet and other state-of-the-art methods.
Besides, KITPose performs better when equipped with HRNet-W48 and obtains $59.1$ mean PCK value.
The results reflect the superiority of the proposed KITPose on AnimalKingdom.
Considering the smaller data volumes and larger data variances of AnimalKingdom, the results confirm the effectiveness of our proposed KIT module in learning the structural-supporting dependencies for general mammal pose estimation.
We observe that similar to the above animal datasets, ViTPose trained only on AnimalKingdom does not perform well either.
Despite CLAMP achieving high performance with a 58.9 PCK@0.05 score, it leverages external textual prior knowledge to guide the pose learning process.
Compared with the current state-of-the-art CLAMP-ViTB, our KITPose-E2C6 only requires single-modality visual information, and outperforms it by 0.2 score.
Interestingly, the recent work, UniAP~\cite{sun2024uniap} proposes a multi-task learning framework to address universal animal perception tasks.
It takes multi-modal labels from the support sets as prompts to guide learning, which achieves impressive performance with 98.59 PCK@0.05 even in the few-shot setting. These results might be ascribed to the prior knowledge provided by support sets, and the use of adaptive weighted loss to enhance knowledge transfer.

%------------------------------ Ablation study of adaptive weight strategy ----------------------------------
\begin{table}[t]
\renewcommand{\arraystretch}{1.2} % set the space between rows
\centering
\setlength\tabcolsep{1.2pt} % set the column space
\resizebox{0.47\textwidth}{!}{
    \begin{tabular}{lllcccc}
    \toprule
    Method & Scheme & AP & AP$^{.50}$ & AP$^{.75}$ & AP$^{M}$ & AP$^{L}$ \\ \midrule\midrule
    \multirow{3}{*}{HRNet-W32}\hspace{0.6em} & Hand-crafted & 73.8  & 95.8  & 80.3  & 52.3  & 74.2 \\
                                & \textbf{Constrained}   & 74.6(\textcolor{red}{$\uparrow0.8$})  & 95.6  & 81.0  & 58.8  & 75.0 \\
                                & \textbf{Adaptive}       & 74.5(\textcolor{red}{$\uparrow0.7$})  & 95.5  & 81.4  & 58.5  & 74.9 \\ \midrule
    \multirow{3}{*}{KITPose-E2}  & Hand-crafted & 75.0 & 95.1 & 81.2 & 57.3 & 75.2  \\
                                & \textbf{Constrained} & 75.6(\textcolor{red}{$\uparrow0.6$}) & 96.0 & \textbf{83.2} & 51.6 & 76.0 \\
                                & \textbf{Adaptive} & \textbf{76.1}(\textcolor{red}{$\uparrow1.1$}) & \textbf{96.6} & 82.8 & \textbf{61.4} & \textbf{76.5} \\
    % Weight Type & AP & AP$^{.50}$ & AP$^{.75}$ & AP$^{M}$ & AP$^{L}$ \\ \midrule\midrule
    % Handcraft factor & 74.7 & 95.8 & 81.3 & 55.1 & 75.1  \\
    % Trainable factor & 75.6 & 96.0 & \textbf{83.2} & 51.6 & 76.0 \\
    % Adaptive map & \textbf{76.1} & \textbf{96.6} & 82.8 & \textbf{61.4} & \textbf{76.5} \\
    \bottomrule
    \end{tabular}
}
% \vspace{1em}
\caption{Results for various keypoint weight strategies for HRNet and KITPose. The input image size is $256\times256$.}
\label{tab:adp_weights}
\end{table}
%----------------------------------------------------------------------------------------------------------
%----------------------------------- Visualization of weight map ------------------------------------------
\begin{figure}[!t]
\centering
\includegraphics[width=\linewidth]{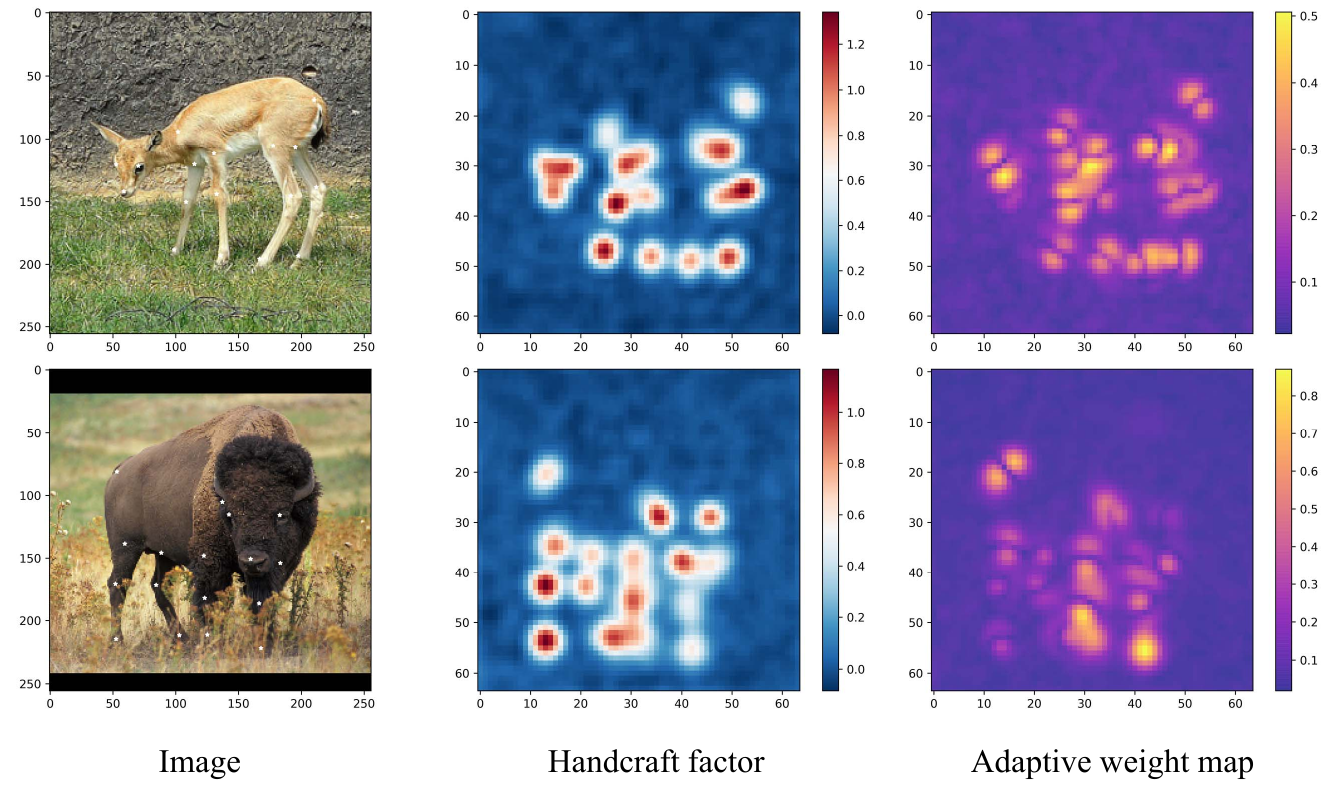}
\caption{Visualisation of adaptive keypoint weight map. From left to right are input images, feature maps, and adaptive weight maps, respectively. We directly multiply the handcraft weight factors with the output feature maps, then sum them up along the channel dimension.}
\label{fig:weight_map}
\end{figure}
%----------------------------------------------------------------------------------------------------------
%------------------------------ Ablation study of adaptive weight strategy ----------------------------------
\begin{table}[t]
\renewcommand{\arraystretch}{1.2} % set the space between rows
\centering
\setlength\tabcolsep{1.2pt} % set the column space
\resizebox{0.48\textwidth}{!}{
    \begin{tabular}{c|@{\hspace{6pt}}c@{\hspace{6pt}}|@{\hspace{6pt}}ccccc}
    \toprule
    Laplacian Kernel    & Input size &   AP  & AP$^{.50}$    & AP$^{.75}$    & AP$^{M}$  & AP$^{L}$ \\ \midrule\midrule
    \multirow{2}{*}{\colvecn{0 & -1 & 0 \\ -1 & 4 & -1 \\ 0 & -1 & 0 \\ }}
        & $256\times 256$ & 76.6 & 96.7 & 84.5 & 57.5 & 76.9 \\
        & $384\times 384$ & 77.2 & 97.3 & 85.0 & 56.4 & 77.6 \\
        \midrule
    \multirow{2}{*}{$\frac14\cdot$\colvecs{0 & 0 & -1 & 0 & 0 \\ 0 & -1 & -2 & -1 & 0 \\ -1 & -2 & 16 & -2 & -1 \\ 0 & -1 & -2 & -1 & 0 \\ 0 & 0 & -1 & 0 & 0 \\}}
    % \hspace{0.6em}
        & $256\times 256$ & 76.8  & 96.1 & 85.1 & 59.2 & 77.2 \\
        & $384\times 384$ & 77.2  & 96.9 & 84.4 & 57.6 & 77.6 \\
        \midrule
    \multirow{2}{*}{$\frac16\cdot$\colvect{0 & 0 & -1 & -1 & -1 & 0 & 0 \\
                0 & -1 & -3 & -3 & -3 & -1 & 0 \\
                -1 & -3 & 0 & 7 & 0 & -3 & -1 \\
                -1 & -3 & 7 & 24 & 7 & -3 & -1 \\
                -1 & -3 & 0 & 7 & 0 & -3 & -1 \\
                0 & -1 & -3 & -3 & -3 & -1 & 0 \\
                0 & 0 & -1 & -1 & -1 & 0 & 0 \\}}
        & $256\times 256$ & 71.5 & 95.1 & 78.1 & 51.7 & 72.0 \\
        & $384\times 384$ & 75.9 & 96.8 & 83.4 & 55.8 & 76.2 \\
    \bottomrule
    \end{tabular}
}
% \vspace{1em}
\caption{Comparisons of different Laplacian kernels at varying input sizes. The Laplacian kernel is employed in GHRL to obtain the satisfied intermediate feature representations.}
\label{tab:laplacian-kernel}
\end{table}
%----------------------------------------------------------------------------------------------------------
%----------------------------------------------- AP10k Qualitative results --------------------------------
\begin{figure*}[t]
\centering
\includegraphics[width=\linewidth]{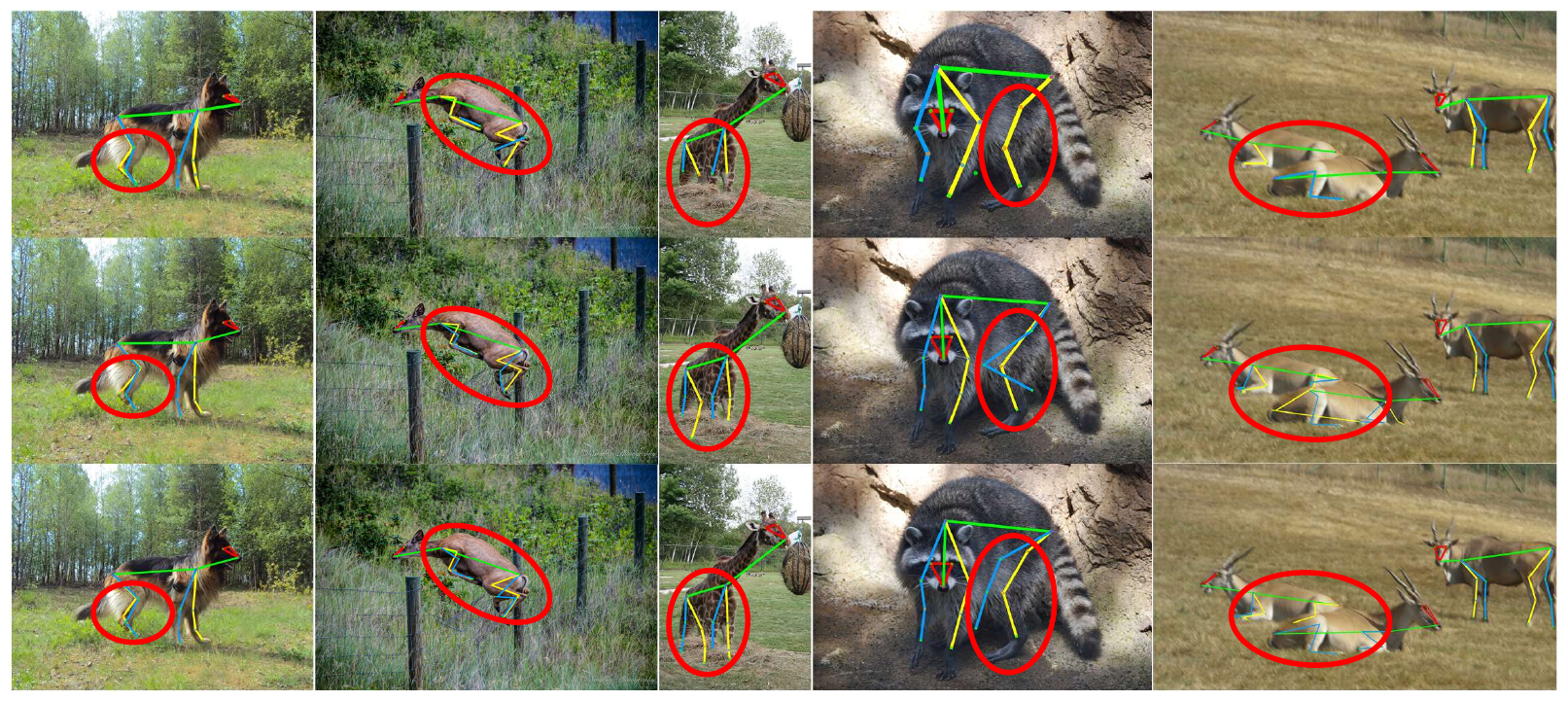}
\caption{Qualitative comparison results of the proposed KITPose (the last row) and the DARK (the second row) on AP10K \textit{val} set. The ground truth poses are shown in the first row.}
\label{fig:quality_vis}
\end{figure*}
%----------------------------------------------------------------------------------------------------------

% ---------------------------------------- COCO Results ---------------------------------------------------
\begin{table*}[]
\renewcommand{\arraystretch}{1.2}
\centering
\begin{tabular}{lccccccccccc}
\toprule Methods     & Backbone    & Input size     & \#Params & GFLOPs & AP   & AP$^{50}$ & AP$^{75}$ & AP$^{\text{M}}$ & AP$^\text{L}$ & AR \\
\midrule\midrule
       % HRNet~\cite{sun2019deep}       & HRNet-W32       & $256\times192$ & 28.5M    & 7.1    & 74.4 & 90.5      & 81.9      & 70.8     & 81.0   & 79.8 \\
       HRNet~\cite{sun2019deep}       & HRNet-W48       & $256\times192$ & 63.6M    & 14.6   & 75.1 & 90.6      & 82.2      & 71.5     & 81.8   & 80.4 \\
       HRNet~\cite{sun2019deep}    & HRNet-W48 & $384\times288$    & 63.6M & 32.9  & 76.3  & 90.8  & 82.9  & 72.3  & 83.4  & 81.2 \\  
       DARK~\cite{zhang2020distribution}   & HRNet-W48 & $256\times192$    & 28.5M & 7.1   & 75.6  & 90.5  & 82.1  & 71.8  & 82.8  & 80.8 \\
       DARK~\cite{zhang2020distribution}   & HRNet-W48 & $384\times288$    & 63.6M & 32.9  & 76.8  & 90.6  & 83.2  & 72.8  & 84.0  & 81.7 \\
       UDP-Pose~\cite{huang2020devil}    & HRNet-W32       & $256\times192$ & 28.7M    & 7.2    & 76.2 & 90.7      & 82.9      &  72.4    & 83.1   & 81.0 \\
       UDP-Pose~\cite{huang2020devil}    & HRNet-W48       & $256\times192$ & 63.8M    & 14.7   & 77.2 & 91.8      & 83.7      & 73.8     & 83.7   & 82.0 \\
       UDP-Pose~\cite{huang2020devil}    & HRNet-W48       & $384\times288$ & 63.8M    & 33.0   & 77.8 & \textbf{92.0}      & 84.3      & 74.2     & 84.5   & 82.5 \\
       % TokenPose~\cite{li2021tokenpose}   & B              & $256\times192$ & 13.5M    & 5.7    & 74.7 & 89.8      & 81.4      & 71.3      & 81.4.  & 80.0 \\
       TokenPose~\cite{li2021tokenpose}   & L/D24          & $256\times192$ & 27.5M    & 11.0   & 75.8 & 90.3      & 82.5      & 72.3      & 82.7   & 80.9 \\
       % TransPose~\cite{yang2021transpose}   & R-A4           & $256\times192$ & 6.0M     & 8.9    & 72.6 & 89.1      & 79.9      & 68.8      & 79.8   & 78.0 \\
       TransPose~\cite{yang2021transpose}   & H-A6           & $256\times192$ & 17.5M    & 21.8   & 75.8 & 90.1      & 82.1      & 71.9      & 82.8   & 80.8 \\
       % HRFormer~\cite{yuan2021hrformer}    & HRFormer-S     & $256\times192$ & 7.8M     & $2.8$  & 74.0 & ${90.2}$  & ${81.2}$  & ${70.4}$ & ${80.7}$ & ${79.4}$ \\
       HRFormer~\cite{yuan2021hrformer}    & HRFormer-B     & $256\times192$ & 43.2M    & $12.2$ & 75.6 & ${90.8}$  & ${82.8}$  & ${71.7}$ & ${82.6}$ & ${80.8}$ \\
       HRFormer~\cite{yuan2021hrformer}    & HRFormer-B    & $384\times288$    & 43.2M & 26.8  & 77.2  & 91.0  & 83.6 & 73.2   & 84.2  & 82.0 \\
       ViTPose-B~\cite{xu2022vitpose}      & ViT-B          & $256\times192$ & 89.9M    & 17.9  & 75.8 & 90.7  & 83.2  & 68.7  & 78.4  & 81.1   \\
       % ViTPose-L~\cite{xu2022vitpose}      & ViT-L          & $256\times192$ & 308.5M    & 59.8  & 78.3 & 91.4  & 85.2  & 74.7  & 85.3  & 83.5   \\
\midrule KITPose-E2C4  & HRNet-W32       & $256\times192$ & 47.4M    & 7.1    & 76.4 & 91.0      & 83.2      & 72.3     & 83.5        & 81.3 \\
       KITPose-E2C4  & HRNet-W48       & $256\times192$ & 80.1M    & 14.6   & 77.2 & 91.1      & 83.7      & 73.3     & 84.2   & 82.0 \\
       KITPose-E2C4   & HRNet-W48 & $384\times288$    & 126.2M    & 32.9 & \textbf{78.4}  & 91.7  & \textbf{84.6}   & \textbf{74.4} & \textbf{85.7}  & \textbf{83.1} \\
       % KITPose-E4  & HRNet-W32       & $256\times192$ & 56.9M    & 7.1    & 76.6 & 91.0      & 83.5      & 72.6     & 83.4   & 81.5 \\
       % KITPose-E4  & HRNet-W48       & $256\times192$ & 91.9M    & 14.6   & \textbf{77.6} & \textbf{91.2}      & \textbf{84.0}      & \textbf{73.7}     & \textbf{84.4}   & \textbf{82.3} \\
\bottomrule
\end{tabular}
\vspace{1em}
\caption{Comparisons with the state-of-the-art methods on COCO \textit{val} set. The number of parameters and GFLOPs are calculated for the pose estimator. The pose estimation methods are measuring w/o using either ensemble models or extra training data. All models are pre-trained on ImageNet.}
\label{tab:coco} 
\end{table*}
%------------------------------------------------------------------------------------------------------------

% --------------------------------------------- COCO test set results ----------------------------------------
\begin{table*}[]
\renewcommand{\arraystretch}{1.2} % set the space between rows
\centering
% \setlength\tabcolsep{1.2pt} % set the column space
% \footnotesize
\begin{tabular}{lccccccccc}
\toprule
Methods      & Source               & Backbone  & Input size    & AP   & AP$^{.50}$ & AP$^{.75}$ & AP$^{\text{M}}$ & AP$^{\text{L}}$ & AR \\
\midrule\midrule
HRNet~\cite{sun2019deep} & CVPR2019   & HRNet-W32 & $256\times192$   & 73.5 & 92.2       & 82.0       & 70.4     & 79.0     & 79.0 \\
HRNet~\cite{sun2019deep} & CVPR2019    & HRNet-W48 & $384\times288$    & 75.5 & 92.5       & 83.3       & 71.9     & 81.5     & 80.5 \\
DARK~\cite{zhang2020distribution} & CVPR2020 & HRNet-W48 & $384\times288$ & 76.2 & 92.5 & 83.6       & 72.5     & 82.4     & 81.1 \\
UDP~\cite{huang2020devil} & CVPR2020   & HRNet-W48 & $384\times288$   & 76.5 & 92.7       & 84.0       & 73.0     & 82.4     & 81.6 \\
CCM~\cite{zhang2021towards} & IJCV2020 & HRNet-W48 & $384\times288$   & 76.6 & 92.8       & 84.1       & 72.6     & 82.4     & 81.7 \\
TokenPose~\cite{li2021tokenpose}   & ICCV2021 & L/D24 & $384\times288$  & 75.9  & 92.3     & 83.4       & 72.2     & 82.1     & 80.8 \\
TransPose~\cite{yang2021transpose} & ICCV2021 & H-A6 & $256\times192$ & 75.0 & 92.2     & 82.3       & 71.3     & 81.1     & 80.1 \\
HRFormer~\cite{yuan2021hrformer}   & NIPS2021 & HRFormer-B & $384\times288$  & 76.2 & 92.7 & 83.8       & 72.5     & 82.3     & 81.2 \\
VitPose-B~\cite{xu2022vitpose} & NIPS2022  & ViT-B & $256\times192$   & 75.1  & 92.5  & 83.1  & 72.0  & 80.7  & 80.3 \\
DGN~\cite{tu2023dual}  & IJCV2023  & HRNet-W48 & $384\times288$    & 75.7  & 92.3  & 83.3 & -  & - & - \\
\midrule
KITPose-E2C4    & Ours  & HRNet-W32 & $256\times192$  & 75.4 & 92.5       & 83.3       & 71.9     & 81.3     & 80.5 \\
KITPose-E2C4    & Ours  & HRNet-W48 & $256\times192$  & 76.4 & 93.0       & 84.1       & 73.0     & 82.3     & 81.4 \\
KITPose-E2C4    & Ours  & HRNet-W48 & $384\times288$  & \textbf{77.3} & \textbf{93.1}       & \textbf{84.7}       & \textbf{73.7}     & \textbf{83.4}     & \textbf{82.2} \\
\bottomrule
\end{tabular}
\caption{Comparison with the SOTA approaches on COCO \textit{test-dev} set. The pose estimation methods are measuring w/o using either ensemble models or extra training data. All models are pre-trained on ImageNet.}
\label{tab:coco_test-dev}
\end{table*}
% ------------------------------------------------------------------------------------------------------------

\subsection{Ablation Study}
\label{subsec:ablation study}
\textbf{Analysis on different components.}
To evaluate the effectiveness of KITPose precisely, we first set experiment groups with different components enabled and all groups are conducted on the AP10K dataset.
The details are shown in Table~\ref{tab:ablation_study}.
Simply applying an adaptive weighting strategy or KIT module can bring relatively small AP gains, at 0.7 and 0.5 respectively.
Combining the two components together can further bring a larger gain of 2.3.
It proves that mining the importance of different keypoints and learning keypoint interactions play a mutually reinforcing role in animal pose estimation.
Note that the body part prompt must be utilised in conjunction with the KIT module.
Adding the body part prompt to the KIT module significantly boosts the AP performance from 73.8 to 75.8.
This demonstrates the effectiveness of our proposed body part prompt, and shows that incorporating more structural-semantic context is helpful for keypoint interactions.
Furthermore, integrating the three components achieves the best performance.
We conclude that learning structural-supporting dependencies, and jointly considering the importance of different keypoints are important to animal pose estimation.

\textbf{Body part prompts Analysis.}
In this part, we first analyse the effect of the number of body part prompts.
Since the protocols of AP10K and AnimalKingdom are quite different (\ie, different keypoint numbers), it is straightforward to expect the analysis of prompt numbers should also differ.
Therefore, we conduct the ablation experiments on the involved animal datasets, the results of AP/AR and mean PCK are shown in Fig.~\ref{fig:cluster-number}.
On AP10K with the 17-keypoint setting, a small prompt number $N_p$ could achieve better performance, \eg KITPose reaches $76.5$ AP score when $N_p=2$.
The best performance ($76.6$ AP score) is obtained when $N_p=4$.
Meanwhile, on AnimalKingdom with the 23-keypoint setting, a larger prompt number ($N_p=6$) achieves the best performance, $59.1$ PCK value.
Through both Fig.~\ref{fig:cluster-sub1} and Fig.~\ref{fig:cluster-sub2}, we also notice that a small prompt number means fewer body part prompts, where the prompts cannot fulfil the objective of assembling features into body parts.
On the contrary, a large prompt number can introduce redundancy.
Therefore, a suitable prompt number can fully activate the advantage of involving body parts.

Table~\ref{tab:bp-params} shows the impact of the clustering-related parameters on experimental performance.
We can observe that with the increase in the number of body part prompts, the speed of clustering slows down.
Meanwhile, other hyper-parameters marginally impact the speed.
Compared to the disabled body part prompt as shown in Table~\ref{tab:ablation_study}, taking the cosine metric leads to worse performance (76.1 vs. 75.8 AP).
Additionally, setting the threshold too high, \ie 1e-2, also results in the same problem, as the model is not sufficiently converged.
The above results indicate that unsupervised clustering is parameter insensitive, and appropriate body part prompts can offer additional structural information to guide the learning process.

\textbf{Analysis in terms of the number of encoder layers.}
In this part, we study the effect of KIT encoder depth.
Specifically, we conduct experiments by varying the number of encoder layers.
As shown in Fig.~\ref{fig:encoder-layers}, the performance grows at the first layer and saturates after adding the second encoder layer.
% Even at the eighth layer, our approach begins to deteriorate.
After that, as the network gets deeper, the performance begins to decrease.
The phenomenon reveals that a shallow KIT is capable of learning structural-supporting dependencies.
Besides, an increasing number of parameters tends to overfit the small animal pose datasets, resulting in inferior performance.
The results indicate that it is sufficient for capturing the structural interactions among keypoints by adding a few extra encoder layers (\ie 2 layers are sufficient).
We argue that the result is partially due to the small volume of animal pose estimation datasets, which also demonstrates that the KIT encoder is very efficient and only consumes marginal computation increase.

\textbf{Analysis on different weighting strategies.}
We design comparative experiments to validate the merits of performing \textit{constrained} and \textit{adaptive} weight strategies respectively.
We use HRNet-W32 and KITPose-E2 as the baseline models with a fixed input size of $256\times256$, and evaluate these models on the AP10K \textit{val} set.
As we can see in Table~\ref{tab:adp_weights}, \textit{constrained manner} can bring an improvement of $+0.8$ AP and $+0.6$ AP for HRNet and KITPose, respectively.
Since the original fixed keypoints weights are sub-optimal, the constrained manner mainly focuses on adjusting them dynamically to change the importance per keypoint.
However, the rough design of the loss function fails to take consideration into instance- and keypoint-specific importance and the imbalance between the foreground (keypoints) and background.
Thus, as shown in Table~\ref{tab:adp_weights}, the \textit{adaptive manner} can further improve the performance, especially for our KITPose with $+1.1$ gain in AP.
In addition, we visualise the different weight maps in Fig.~\ref{fig:weight_map}.
We have marked the locations of the keypoints with a star-shaped character $\star$ in the input images.
In particular, we directly multiply the handcraft weight factors with the output feature maps, then sum up values along the channel dimension.
% The colourmaps are drawn in red style, \ie the redder a pixel is, the larger its value corresponds.
Therefore, as we can see in these two examples, the adaptive weight map dynamically assigns distinct weights to different keypoints according to different instances.
% heavily concentrates the loss on hard samples, effectively down-weight the effect of easy samples.
Compared to handcraft weight factors, the adaptive weight map performs quite differently from the handcraft weight factor.
In detail, for bison in Fig.~\ref{fig:weight_map}, the adaptive weight map focuses much attention on the left front paw and the root of the tail, while handcraft weight factors produce the opposite.
Following this way, the adaptive weight strategy is indispensable for addressing the imbalance problem in pose estimation, further enhancing our structural-supporting solution.
Furthermore, combining it with body part prompt yields even more significant improvements (\ie 76.1 to 76.6 AP).

\textbf{Analysis on Laplacian kernel.}
In general, the Laplacian is a high-pass filter that amplifies high-frequency signals. 
Thus, the Laplacian kernel plays a crucial role in our proposed GHRL, where it is utilised to control the sharpness of the Gaussian target.
The results with 3 kernel settings are listed in Table~\ref{tab:laplacian-kernel}.
From top to bottom in the table, the kernel sizes are $3\times 3$, $5\times 5$, and $7\times 7$, respectively.
In order to ensure the consistency of the Gaussian peak value, it is necessary to multiply by factors of $\frac14$ and $\frac16$ for $5\times 5$ and $7\times 7$ kernels.
We can see that $5\times 5$ Laplacian kernel consistently achieves the best results 76.8 and 77.2 AP, regardless of the input resolution.
As the kernel size increases, a degradation in performance is observed.
We analyse the reason lies in the conflict between the size of the Laplacian kernel and Gaussian target (6 in $256\times 256$ and 9 in $384\times 384$).
When the Laplacian kernel reaches or even exceeds the size of the Gaussian target, it suppresses the supervision signal and sacrifices the localisation power.
Since the Laplacian is a second-order derivative operator, its ability to pinpoint the exact location of the Gaussian edge compromises when the kernel size is too large.
% Meanwhile, the kernel's sensitivity to changes in intensity, which correspond to Gaussian target, is reduced.
% This means that the GHRL may fail to accurately localise keypoint.
The above analysis illustrates the rationale behind the adoption of Laplacian kernels in GHRL, and also elucidates their relative stability with different input sizes.

%------------------------------------- COCO Qualitative results ----------------------------------------------------
\begin{figure*}[]
    \centering
    \includegraphics[width=\linewidth]{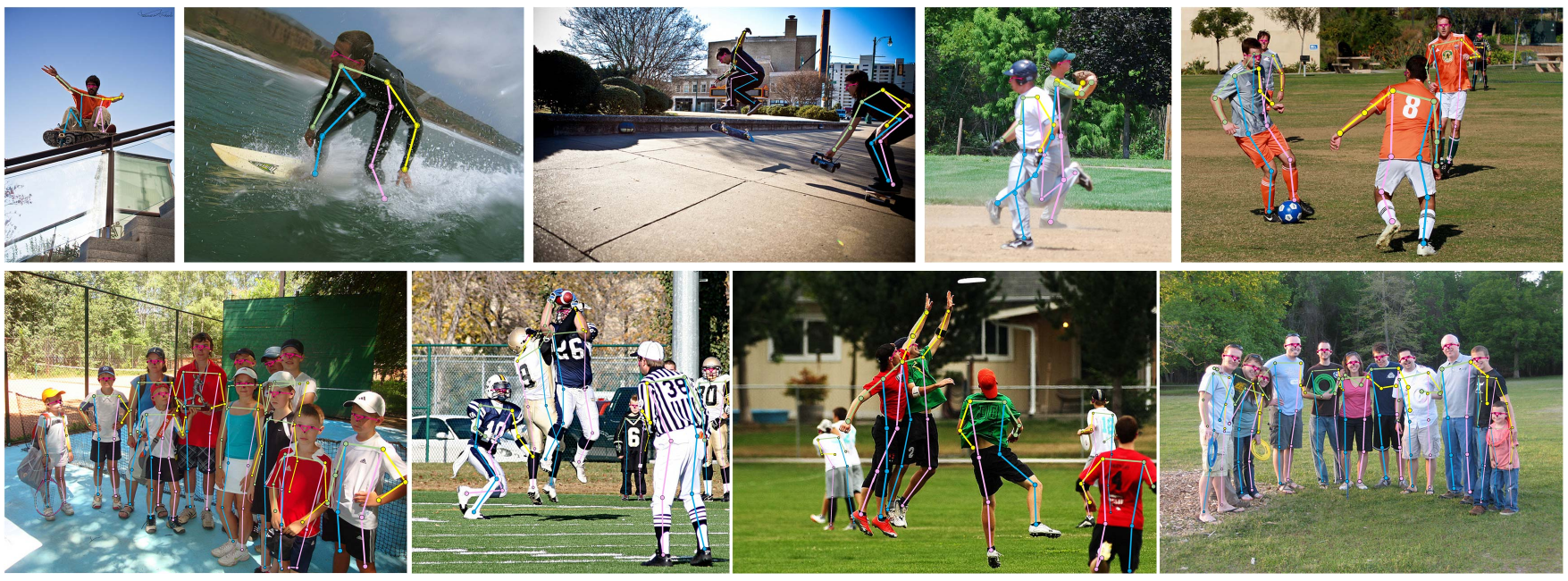}
    \caption{Qualitative results of KITPose on COCO pose estimation validation set, including different viewpoints, changing appearances, multiple targets, and occlusion.}
    \label{fig:coco_quality_results}
\end{figure*}
%------------------------------------------------------------------------------------------------------------------

\subsection{Generalisation to Human Pose Estimation}
\textbf{Dataset \& Evaluation metric.}
To evaluate the generalisation performance of KITPose for non-animal mammal (\ie human) pose estimation, we extend to evaluate the MS COCO~\cite{lin2014microsoft} dataset.
The MS COCO dataset contains real-world samples of humans, including the structure of their poses.
% It is the typical dataset in the field of human pose estimation.
There are over $200K$ images and $250K$ instances with $17$ labelled keypoints in the COCO dataset.
We use the standard evaluation metric, Average Precision (AP) based on Object Keypoint Similarity (OKS), to measure the performance.
Mean AP and APs at different OKS thresholds are reported, \ie AP, AP.5 and AP.75, as well as APs for different instance sizes,\ie AP$^{\text{M}}$ and AP$^{\text{L}}$.
Besides, mean Average Recall (AR) is also reported in our experiments.
We follow the standard train2017/val2017/test-dev2017 splits for the evaluation, containing 118K, 5K and 20K images, respectively.

\textbf{Training.}
Following the conventional setting of~\cite{sun2019deep}, we augment the images by random rotation ($-45^\circ\sim45^\circ$), random scaling ($0.65\sim1.35$), and half body transformation.
Apart from the above settings, we apply cutmix to enhance model inferring capability.
In our experiments, we evaluate with the input resolutions being $256\times192$ and $384\times288$.
All the HRNet backbones and KIT modules are optimised by ADAM with a base learning rate of $5$e-$4$.
Each model is trained for $250$ epochs total and the learning rate is decreased by $0.1$ times at the $200$th and $230$th epochs.

\textbf{Testing.}
For a fair comparison, we use the same person detector provided by Simple Baseline~\cite{xiao2018simple} for both validation and testing.
According to the bounding box generated by the person detector, the single-person image is cropped from the original image and resized to a fixed resolution, \eg $384\times288$.
During testing, we follow the random horizontal-flip method and average the original and flipped images.

\textbf{Comparison with the state-of-the-art methods.}
In Table~\ref{tab:coco}, as we can see, our proposed KITPose achieves superior performance compared with the previous state-of-the-art methods on the COCO \textit{val2017} set.
After connecting our KIT module to HRNet-W48 backbones, we observe that our KITPose-E2C4 improves $+0.8$ AP at an input size of $256\times256$ (KITPose $76.4$ vs. DarkPose-W32 $75.6$), which verifies the effectiveness of our proposed method.
Although our KITPose performs equivalently to the SOTA UDP-Pose at input resolution $256\times192$, we outperform UDP-Pose by $+0.6$ AP at an input resolution of $384\times288$.
We also outperform ViTPose-B by $+1.4$ AP with comparable model parameters and computational complexity (KITPose $77.2$ vs. ViTPose-B $75.8$).
Besides, Table~\ref{tab:coco_test-dev} reports the performance of our method and the existing state-of-the-art methods on the COCO \textit{test-dev2017} set.
The proposed KITPose method, trained based on HRNet-W48 with the input size $256\times192$, achieves a $76.4$ AP score, outperforming other SOTA methods with the same input size, with even higher accuracy than other Transformer-based models with the input size $384\times288$ (KITPose $76.4$ vs. HRFormer $76.2$ and ViTPose $75.1$).
With a bigger input size $384\times288$, our KITPose achieves the highest $77.3$ AP without any sophisticated strategies and extra training data.
The above results and analysis demonstrate the merit of learning structure-supporting dependencies in advancing general pose estimation tasks, for constructing both human and animal pose estimation solutions.
% When using essentially the same data augmentation as HRNet except for the use of cutmix, we improve $+2.5$ AP points.
% Under the same input size, our KITPose-E2 and KITPose-E4 have outperformed the SOTA HRFormer by $+0.5$ AP and $+2.0$ AP, respectively.

\section{Conclusion}
In this work, to achieve general mammal pose estimation, we developed a novel keypoints-interactive model, namely KITPose, to pursue structure-supporting dependencies among keypoints and body parts.
In particular, we explicitly disentangle the keypoint-specific clues from the backbone features without any spatial splitting.
% The intermediate keypoint heatmaps supervision is optimised with a novel Generalised Heatmap Regression Loss.
An effective design named Generalised Heatmap Regression Loss is proposed to enable the adaptive adjustment of intermediate features to optimise keypoint representations.
Simultaneously, to preserve the semantic information in the image, a new concept, referred to as body part prompts, is introduced to provide discriminative context, organising the information interactions.
Furthermore, to automatically balance the importance between each keypoints, a novel adaptive weight strategy is introduced to common MSE loss.
The designed architecture reflects its superiority and generalisation for general mammal pose estimation, which has been evaluated through extensive experiments on the AP10K, AnimalKingdom, and COCO datasets.

\section*{Availability of Data and Materials}
The datasets generated during and/or analyzed during the current study are available from the corresponding author on reasonable request.
\section*{Competing Interests}
The authors have no competing interests to declare that are relevant to the content of this article.

\begin{acknowledgements}
This work was supported in part by the National Natural Science Foundation of China (62106089, 62020106012, 62332008, 62336004).
\end{acknowledgements}


% \bibliographystyle{spbasic}       % basic style, author-year citations
% \bibliography{ref.bib}

\begin{thebibliography}{48}
\providecommand{\natexlab}[1]{#1}
\providecommand{\url}[1]{{#1}}
\providecommand{\urlprefix}{URL }
\expandafter\ifx\csname urlstyle\endcsname\relax
  \providecommand{\doi}[1]{DOI~\discretionary{}{}{}#1}\else
  \providecommand{\doi}{DOI~\discretionary{}{}{}\begingroup \urlstyle{rm}\Url}\fi
\providecommand{\eprint}[2][]{\url{#2}}

\bibitem[1]{cao2019cross}
Cao J, Tang H, Fang HS, Shen X, Lu C, Tai YW (2019) Cross-domain adaptation for animal pose estimation. In: IEEE International Conference on Computer Vision, pp 9498--9507

\bibitem[2]{mmpose2020}
Contributors M (2020) Openmmlab pose estimation toolbox and benchmark. \url{https://github.com/open-mmlab/mmpose}

\bibitem[3]{davies2015keep}
Davies K (2015) Keep the directive that protects research animals. Nature 521(7550):7--7

\bibitem[4]{deng2009imagenet}
Deng J, Dong W, Socher R, Li LJ, Li K, Fei-Fei L (2009) Imagenet: A large-scale hierarchical image database. In: IEEE Computer Vision and Pattern Recognition, pp 248--255

\bibitem[5]{dosovitskiy2020image}
Dosovitskiy A, Beyer L, Kolesnikov A, Weissenborn D, Zhai X, Unterthiner T, Dehghani M, Minderer M, Heigold G, Gelly S, et~al. (2020) An image is worth 16x16 words: Transformers for image recognition at scale. arXiv preprint arXiv:201011929

\bibitem[6]{hirschorn2023pose}
Hirschorn O, Avidan S (2023) Pose anything: A graph-based approach for category-agnostic pose estimation. arXiv preprint arXiv:231117891

\bibitem[7]{huang2020devil}
Huang J, Zhu Z, Guo F, Huang G (2020) The devil is in the details: Delving into unbiased data processing for human pose estimation. In: IEEE Computer Vision and Pattern Recognition, pp 5700--5709

\bibitem[8]{jiang2022animal}
Jiang L, Lee C, Teotia D, Ostadabbas S (2022) Animal pose estimation: A closer look at the state-of-the-art, existing gaps and opportunities. Computer Vision and Image Understanding, p 103483

\bibitem[9]{katsavounidis1994new}
Katsavounidis I, Kuo CCJ, Zhang Z (1994) A new initialization technique for generalized lloyd iteration. IEEE Signal Processing Letters 1(10):144--146

\bibitem[10]{ke2018multi}
Ke L, Chang MC, Qi H, Lyu S (2018) Multi-scale structure-aware network for human pose estimation. In: European Conference on Computer Vision, pp 713--728

\bibitem[11]{kingma2014adam}
Kingma DP, Ba J (2014) Adam: A method for stochastic optimization. arXiv preprint arXiv:14126980

\bibitem[12]{lauer2022multi}
Lauer J, Zhou M, Ye S, Menegas W, Schneider S, Nath T, Rahman MM, Di~Santo V, Soberanes D, Feng G, et~al. (2022) Multi-animal pose estimation, identification and tracking with deeplabcut. Nature Methods 19(4):496--504

\bibitem[13]{li2021synthetic}
Li C, Lee GH (2021) From synthetic to real: Unsupervised domain adaptation for animal pose estimation. In: IEEE Computer Vision and Pattern Recognition, pp 1482--1491

\bibitem[14]{li2023scarcenet}
Li C, Lee GH (2023) Scarcenet: Animal pose estimation with scarce annotations. In: IEEE Computer Vision and Pattern Recognition, pp 17174--17183

\bibitem[15]{li2021prtr}
Li K, Wang S, Zhang X, Xu Y, Xu W, Tu Z (2021{\natexlab{a}}) Pose recognition with cascade transformers. In: IEEE Computer Vision and Pattern Recognition, pp 1944--1953

\bibitem[16]{li2020generalized}
Li X, Wang W, Wu L, Chen S, Hu X, Li J, Tang J, Yang J (2020) Generalized focal loss: Learning qualified and distributed bounding boxes for dense object detection. Advances in Neural Information Processing Systems 33:21002--21012

\bibitem[17]{li2021tokenpose}
Li Y, Zhang S, Wang Z, Yang S, Yang W, Xia ST, Zhou E (2021{\natexlab{b}}) Tokenpose: Learning keypoint tokens for human pose estimation. In: IEEE International Conference on Computer Vision, pp 11313--11322

\bibitem[18]{lin2014microsoft}
Lin TY, Maire M, Belongie S, Hays J, Perona P, Ramanan D, Doll{\'a}r P, Zitnick CL (2014) Microsoft coco: Common objects in context. In: European Conference on Computer Vision, pp 740--755

\bibitem[19]{lin2017focal}
Lin TY, Goyal P, Girshick R, He K, Doll{\'a}r P (2017) Focal loss for dense object detection. In: IEEE International Conference on Computer Vision, pp 2980--2988

\bibitem[20]{liu2021swin}
Liu Z, Lin Y, Cao Y, Hu H, Wei Y, Zhang Z, Lin S, Guo B (2021) Swin transformer: Hierarchical vision transformer using shifted windows. In: IEEE International Conference on Computer Vision, pp 10012--10022

\bibitem[21]{mao2022towards}
Mao X, Qi G, Chen Y, Li X, Duan R, Ye S, He Y, Xue H (2022) Towards robust vision transformer. In: IEEE Computer Vision and Pattern Recognition, pp 12042--12051

\bibitem[22]{mathis2018deeplabcut}
Mathis A, Mamidanna P, Cury KM, Abe T, Murthy VN, Mathis MW, Bethge M (2018) Deeplabcut: markerless pose estimation of user-defined body parts with deep learning. Nature Neuroscience 21(9):1281--1289

\bibitem[23]{mathis2021pretraining}
Mathis A, Biasi T, Schneider S, Yuksekgonul M, Rogers B, Bethge M, Mathis MW (2021) Pretraining boosts out-of-domain robustness for pose estimation. In: Winter Conference on Applications of Computer Vision, pp 1859--1868

\bibitem[24]{mu2020learning}
Mu J, Qiu W, Hager GD, Yuille AL (2020) Learning from synthetic animals. In: IEEE Computer Vision and Pattern Recognition, pp 12386--12395

\bibitem[25]{naseer2021intriguing}
Naseer MM, Ranasinghe K, Khan SH, Hayat M, Shahbaz~Khan F, Yang MH (2021) Intriguing properties of vision transformers. Advances in Neural Information Processing Systems 34:23296--23308

\bibitem[26]{ng2022animal}
Ng XL, Ong KE, Zheng Q, Ni Y, Yeo SY, Liu J (2022) Animal kingdom: A large and diverse dataset for animal behavior understanding. In: IEEE Computer Vision and Pattern Recognition, pp 19023--19034

\bibitem[27]{park2022how}
Park N, Kim S (2022) How do vision transformers work? In: International Conference on Learning Representations

\bibitem[28]{pereira2019fast}
Pereira TD, Aldarondo DE, Willmore L, Kislin M, Wang SSH, Murthy M, Shaevitz JW (2019) Fast animal pose estimation using deep neural networks. Nature Methods 16(1):117--125

\bibitem[29]{pereira2022sleap}
Pereira TD, Tabris N, Matsliah A, Turner DM, Li J, Ravindranath S, Papadoyannis ES, Normand E, Deutsch DS, Wang ZY, et~al. (2022) Sleap: A deep learning system for multi-animal pose tracking. Nature Methods 19(4):486--495

\bibitem[30]{radford2021learning}
Radford A, Kim JW, Hallacy C, Ramesh A, Goh G, Agarwal S, Sastry G, Askell A, Mishkin P, Clark J, et~al. (2021) Learning transferable visual models from natural language supervision. In: International Conference on Machine Learning, pp 8748--8763

\bibitem[31]{rao2022kitpose}
Rao J, Xu T, Song X, Feng ZH, Wu XJ (2022) Kitpose: Keypoint-interactive transformer for animal pose estimation. In: Chinese Conference on Pattern Recognition and Computer Vision, Springer, pp 660--673

\bibitem[32]{sun2019deep}
Sun K, Xiao B, Liu D, Wang J (2019) Deep high-resolution representation learning for human pose estimation. In: IEEE Computer Vision and Pattern Recognition, pp 5693--5703

\bibitem[33]{sun2024uniap}
Sun M, Zhao Z, Chai W, Luo H, Cao S, Zhang Y, Hwang JN, Wang G (2024) Uniap: Towards universal animal perception in vision via few-shot learning. In: Proceedings of the AAAI Conference on Artificial Intelligence, vol~38, pp 5008--5016

\bibitem[34]{tu2023dual}
Tu J, Wu G, Wang L (2023) Dual graph networks for pose estimation in crowded scenes. International Journal of Computer Vision pp 1--21

\bibitem[35]{xiao2018simple}
Xiao B, Wu H, Wei Y (2018) Simple baselines for human pose estimation and tracking. In: European Conference on Computer Vision, pp 466--481

\bibitem[36]{xu2022pose}
Xu L, Jin S, Zeng W, Liu W, Qian C, Ouyang W, Luo P, Wang X (2022{\natexlab{a}}) Pose for everything: Towards category-agnostic pose estimation. In: European conference on computer vision, pp 398--416

\bibitem[37]{xu2022vitpose}
Xu Y, Zhang J, Zhang Q, Tao D (2022{\natexlab{b}}) Vi{TP}ose: Simple vision transformer baselines for human pose estimation. In: Advances in Neural Information Processing Systems

\bibitem[38]{xu2022vitpose+}
Xu Y, Zhang J, Zhang Q, Tao D (2022{\natexlab{c}}) Vitpose+: Vision transformer foundation model for generic body pose estimation. arXiv preprint arXiv:221204246

\bibitem[39]{yang2021transpose}
Yang S, Quan Z, Nie M, Yang W (2021) Transpose: Keypoint localization via transformer. In: IEEE International Conference on Computer Vision, pp 11802--11812

\bibitem[40]{yu2021ap}
Yu H, Xu Y, Zhang J, Zhao W, Guan Z, Tao D (2021) Ap-10k: A benchmark for animal pose estimation in the wild. In: Advances in Neural Information Processing Systems

\bibitem[41]{yuan2021hrformer}
Yuan Y, Fu R, Huang L, Lin W, Zhang C, Chen X, Wang J (2021) Hrformer: High-resolution vision transformer for dense predict. Advances in Neural Information Processing Systems 34:7281--7293

\bibitem[42]{yun2019cutmix}
Yun S, Han D, Oh SJ, Chun S, Choe J, Yoo Y (2019) Cutmix: Regularization strategy to train strong classifiers with localizable features. In: IEEE International Conference on Computer Vision, pp 6023--6032

\bibitem[43]{zhang2020distribution}
Zhang F, Zhu X, Dai H, Ye M, Zhu C (2020) Distribution-aware coordinate representation for human pose estimation. In: IEEE Computer Vision and Pattern Recognition, pp 7093--7102

\bibitem[44]{zhang2021towards}
Zhang J, Chen Z, Tao D (2021) Towards high performance human keypoint detection. International Journal of Computer Vision 129(9):2639--2662

\bibitem[45]{zhang2023animaltrack}
Zhang L, Gao J, Xiao Z, Fan H (2023{\natexlab{a}}) Animaltrack: A benchmark for multi-animal tracking in the wild. International Journal of Computer Vision 131(2):496--513

\bibitem[46]{zhang2023clamp}
Zhang X, Wang W, Chen Z, Xu Y, Zhang J, Tao D (2023{\natexlab{b}}) Clamp: Prompt-based contrastive learning for connecting language and animal pose. In: IEEE Computer Vision and Pattern Recognition, pp 23272--23281

\bibitem[47]{zhao2022centerclip}
Zhao S, Zhu L, Wang X, Yang Y (2022) Centerclip: Token clustering for efficient text-video retrieval. In: International ACM SIGIR Conference on Research and Development in Information Retrieval, pp 970--981

\bibitem[48]{zuffi2019three}
Zuffi S, Kanazawa A, Berger-Wolf T, Black MJ (2019) Three-d safari: Learning to estimate zebra pose, shape, and texture from images" in the wild". In: IEEE Computer Vision and Pattern Recognition, pp 5359--5368

\end{thebibliography}

% % Non-BibTeX users please use
% \begin{thebibliography}{}
% %
% % and use \bibitem to create references. Consult the Instructions
% % for authors for reference list style.
% % Format for Journal Reference
% Author, Article title, Journal, Volume, page numbers (year)
% Author, Book title, page numbers. Publisher, place (year)
% \end{thebibliography}

\end{document}